\documentclass{CUP-JNL-NLP}

\usepackage{graphicx}
\usepackage{natbib}
\ifpdf%
\usepackage{epstopdf}%
\else%
\fi
\usepackage{multirow}

\usepackage{amsmath}
\usepackage{booktabs}
\usepackage{array}
\usepackage{comment}
\usepackage{multicol}
\usepackage{multirow}
\usepackage{pifont}
\usepackage{enumerate}
\usepackage{xcolor, colortbl}
\usepackage[dvipsnames]{xcolor}
\usepackage{worldflags}
\usepackage{longtable}
\usepackage{subcaption}
\usepackage{url}

\definecolor{OliveGreen}{HTML}{60A917}

\newcommand{\cmark}{\ding{51}}%
\newcommand{\xmark}{\ding{55}}%

\newcommand{\yt}{\cellcolor{OliveGreen!100}\color{white}\vspace{-0.3cm}\scriptsize\cmark}
\newcommand{\nt}{\cellcolor{BrickRed!100}\color{white}\vspace{-0.3cm}\scriptsize\xmark}
\newcommand{\mt}{\cellcolor{Gray!25}\color{black}\vspace{-0.3cm}\large\textbf{$\sim$}}

\begin{document}
\label{firstpage}

\lefttitle{}
\righttitle{}

\papertitle{}

\jnlPage{\pageref{firstpage}}{\pageref{lastpage}}
\jnlDoiYr{}

\title{Language Models for Portuguese: A Systematic Mapping Study}

\begin{authgrp}

\author{Jhessica Silva}
\affiliation{Instituto de Computação, Universidade Estadual de Campinas (UNICAMP), Campinas, 13083-852, São Paulo, Brazil
\email{\{jhessica.silva, sandra, helio\}@ic.unicamp.br}}

\author{Carlos Caetano}

\author{Helena Maia}

\author{Breno Bernard Nicolau de França}

\author{Sandra Avila}

\author{~Helio Pedrini}

\end{authgrp}


\begin{abstract}
In recent years, the rapid development of language models has transformed the field of Natural Language Processing through a wide range of applications. However, the development of language models has not progressed uniformly across all languages. In the case of the Portuguese language, there has recently been a growing effort by academia and companies to develop language models and create data resources for Portuguese. These efforts have resulted in the rise of an increasingly diverse ecosystem of language models for Portuguese. However, information on these models remains dispersed in scientific publications, technical reports, model repositories, and project documentation. This survey presents a systematic mapping study of language models developed for Portuguese, providing a comprehensive overview of the current state of the field. We map a total of 46 models, characterizing them by various aspects, including base model, architecture, computational resources, training datasets, licensing, code availability, data, and model weights. Furthermore, we analyzed the evolution and relationships among these models through a phylogenetic perspective, identified current research gaps and opportunities, and discussed future directions for the development of language models for Portuguese.
\end{abstract}

\maketitle

\section{Introduction}

Large Language Models (LLMs) have become one of the main advances on Artificial Intelligence (AI) recently and have significantly transformed the field of Natural Language Processing (NLP)~\citep{NatureComputationalScience:2025}. Such models learn statistical patterns from large collections of textual data and become capable of understanding, generating and manipulating data allowing interaction with users more naturally. Their rapid development has led to a wide range of applications in academia~\citep{Zhang:2025:llms4all} and industry~\citep{Raza:Nature:2025}.

The evolution from early language models to LLMs has further expanded these applications, driven by advances in model architectures, computational resources and the availability of large-scale training corpora~\citep{Raeini:2025:NLPJ}. Today, LLMs have become a fundamental component on modern AI systems supporting a broad spectrum of applications, including machine translation~\citep{Zhu:2024:multilingual}, question answering~\citep{Zhu:2025:QuestionAnswering}, text summarization~\citep{Liu:2024:Summarization}, conversational assistants~\citep{Dhavalikar:2025:Conversational}, code generation~\citep{Jiang:2026:CodeGeneration}, information retrieval~\citep{Zhu:2025:Retrieval}, healthcare~\citep{Vishwanath:2026:Healthcare}, and education~\citep{Oh:2026:Educational}.

Despite these advances, the development of LLMs has not progresses uniformly across all languages~\citep{Li:2025:ExtLowResource, Huang:2026:Multilingualism}. The availability of large-scale textual corpora, computational power and benchmarks has concentrated research efforts on so called \textit{high-resource languages} (e.g., English and Chinese), which benefit from abundant digital content as well as established NLP advances~\citep{Lupascu:2026:LowResourceSurvey}. In contrast, according to~\citet{Das:2024:ColonialImpulse}, many other languages have historically received less attention due to the limited availability of such resources, making the development of language models more challenging. Although Portuguese has traditionally been considered less resourced than English, recent years have witnessed a growing effort from both academia and industry to develop language models and create new linguistic resources tailored to the Portuguese language~\citep{silva_govbert-br_2025, cruz-castaneda_amadeus-verbo_2025, zago_bertugues_2024, correa_tucano_2024, dos_santos_capivara_2023, petrobert, albertina, schneider_gpt-2_2021, bertau, souza_bertimbau_2020, pierre2020gpt2smallportuguese}.

These efforts have resulted in the emergence of an increasingly diverse ecosystem of language models for Portuguese. Since 2020, numerous models have been proposed for Brazilian Portuguese and European Portuguese, ranging from general purpose tasks~\citep{pierre2020gpt2smallportuguese, pires_sabia_2023, miquelina_generating_2022, garcia_gembode_2025, cruz-castaneda_amadeus-verbo_2025} to domain-specific systems~\citep{devlin2019bert, schneider_gpt-2_2021, petrobert, viegas_jurisbert_2023, de_souza_pinto_developing_2025}. Such models are developed by both academia and industry, and adopt different architectures, training strategies, datasets and target domains.

Beyond the academic community, governments have also recognized the strategic importance of developing AI technologies tailored to the Portuguese language. Examples include the Brazilian Artificial Intelligence Plan (\textit{Plano Brasileiro de Inteligência Artificial} -- PBIA)~\citep{pbia} and Portugal's National Artificial Intelligence Agenda (\textit{Agenda Nacional de Inteligência Artificial} -- ANIA)~\citep{ania}. Both initiatives recognize AI capabilities, infrastructure, and language technologies as strategic priorities for their countries. These initiatives further reinforce the importance of understanding the current state of  language models for Portuguese to support future research, technological development, and evidence-based decision-making.

Although many models have been proposed, the current scenario of  language models for Portuguese remains fragmented. Information about the existing models is dispersed across scientific publications, technical reports, model repositories, and project documentation. As a result, it is difficult to obtain a comprehensive understanding of the current state of the field. Moreover, important aspects such as model availability, training data, openness, documentation quality, architectural evolution, and current research gaps have not yet been systematically characterized. In view of that, this scenario highlights the need for a comprehensive systematic mapping study to consolidate the current state of the art and provide a broader understanding of language models developed for the Portuguese language.

To address the aforementioned challenges, this paper presents a systematic mapping study of language models developed for the Portuguese language. Following a rigorous review methodology complemented by the snowballing process~\citep{wohlin2014guidelines}, we identify and analyze language models published between 2020 and 2025, providing a comprehensive overview of the current state of the field.

In addition to identifying the existing language models, this study characterizes them according to multiple dimensions, including base model, architecture, computational resources, training datasets, licensing, code availability, data and model weights. Furthermore, we analyze the evolution and relationships among these models through a phylogenetic perspective, identify current research gaps and opportunities, and discuss future directions for the development of language models for Portuguese.

The remainder of this paper is organized as follows. Section~\ref{sec:related} reviews the most relevant studies on  language models for Portuguese and positions our work with respect to the existing literature. Section~\ref{sec:method} describes the systematic mapping methodology adopted in this study. Section~\ref{sec:findings} presents the identified language models and the results of the mapping. Section~\ref{sec:ad} analyzes and discusses the findings. Section~\ref{sec:gaps} identifies the main research gaps and future opportunities for  language models for Portuguese. Finally, Section~\ref{sec:conclusion} concludes the paper.

\section{Related Work}
\label{sec:related}

This section presents the papers most relevant to this systematic mapping study that address language models for Portuguese.  We provide an overview of them and highlight how they differ from our work.

\citet{correa_tucano_2024} propose a new language model for Brazilian Portuguese, Tucano, and demonstrate the advances of this new model over existing ones. The authors surveyed 24 language models for Portuguese released between 2020 and October 2024, providing brief descriptions of each model and distinguishing between Brazilian Portuguese (PT-BR) and European Portuguese (PT-PT) models. In~\citep{tucano2}, the authors extend their previous work by proposing the Tucano~2 model, and in their new paper, they mapped five new language models~\citep{almeida2025curioedu7bexaminingdata, gaia-gemma-3-4b-2025, jurema, gamallo_galician-portuguese_2025, cruz-castaneda_amadeus-verbo_2025} for Portuguese published in 2025, to compare them to the new model they were launching for Brazilian Portuguese.

\citet{Assis_Freitas_Paes_2025} analyzed six large language models for Brazilian Portuguese~\citep{correa_tucano_2024, wandgibautperiquito3B, garcia_introducing_2024, botbotroboticscabra_2024, luciano_santa_brigida_2024, abonizio_sabia-3_2025} to assess their generative performance on Natural Language Generation tasks: text summarization, text simplification, and generative question answering, comparing the results of these Brazilian Portuguese models with GPT-4o~\citep{hurst2024gpt}. Although generative analysis was the main objective of their paper, the authors, in selecting the six models, conducted a survey of large language models for Brazilian Portuguese via the Open Portuguese LLM Leaderboard\footnote{\url{https://huggingface.co/spaces/eduagarcia/open_pt_llm_leaderboard}} available on Hugging Face, thus presenting a timeline of 29 models released between 2020 \mbox{and 2024}.

\citet{Cruz-Castaneda_Amadeus_2025} conducted a chronological survey of large language models for Brazilian Portuguese released between 2020 and March 2025, comparing these models by architecture (whether encoder-decoder, encoder-only, or decoder-only), number of parameters, and computational resources required for development, such as energy consumption. In total, the authors mapped 45 models, including 16 models that our study did not cover, because our methodology does not consider unpublished models, nor models proposed to solve only a specific task (Exclusion criteria - E3, see Section~\ref{sec:IE})~\citep{souza2020portuguesenamedentityrecognition, pellicer_ptt5-paraphraser_2022, zanuz_fostering_2022, alcoforado_zeroberto_2022, jose_mrat-sqlgap_2021, bonifacio_study_2020, mambarim, boana, nicholas22aira, gaia-gemma-3-4b-2025, Caramelo, Harpia, Caramelinho, Caramelinho, aplptbr, ciurlino2021bertbr, RobertaTwitterBR}.

Unlike prior work, we conducted a systematic mapping study following a rigorous, transparent, and replicable method, double-checked~\citep{kitchenham2015evidence}. We present a mapping of language models for Portuguese, regardless of variant, published between 2020 and August 2025, discussing them in terms of the quality of the article or technical report, the model’s availability in terms of data, code, and weights, the data used in training, and the phylogenetic tree of these models, thereby discussing the architectures used and the necessary computational resources. Finally, we discussed the main gaps and opportunities in language models for Portuguese. Other papers have also addressed language models involving the Portuguese language, but with different specific focuses, such as low-resource languages~\citep{LUPASCU2026104189}, European languages~\citep{ali2024surveylargelanguagemodels}, or healthcare~\citep{shimaoka2026large}, and thus did not address the same objectives of our systematic mapping study.

\section{Research Method}
\label{sec:method}

The goal of this study is to characterize language models developed on Portuguese-language textual data. Models released from 2020 onward are desired that can perform any activity, as long as it focuses on Portuguese, and can be unimodal or multimodal. To achieve this, we conducted a systematic mapping study of the literature on language models for Portuguese. 

Considering the presented goal, this study aims to answer the following research questions:
\begin{enumerate}[\hspace{0.3cm}Q1.]
    \item What language models have been proposed for the Portuguese language?
    \item What is the availability --- in terms of model, code, and data --- of these models?
    \item Which benchmarks or datasets were these models trained on?
    \item Is the training data originate in Portuguese, or is it translated from another language, or artificially generated?
\end{enumerate}  

\subsection{Search Process and Strategy}
\label{sec:peb}

For this study,  we conducted an automated search in the main libraries used for publication of language models papers: Scopus, IEEEXplore, and Web of Science. The automated search was also conducted on arXiv, as many language models are first published there, and some are never officially published in conferences or journals in the field, since they are not papers but technical reports on the proposed model. Also, we complemented this automated search with forward and backward snowballing process~\citep{wohlin2014guidelines}. To calibrate the candidate search string, five primary studies were used as control papers:

\begin{enumerate}[\hspace{0.3cm}1.]
    \item BERTabaporu: Assessing a Genre-Specific Language Model for Portuguese NLP~\citep{da_costa_bertabaporu_2023}
    \item BERTweet.BR: A pre-trained language model for tweets in Portuguese~\citep{carneiro_bertweetbr_2025}
    \item CAPIVARA: Cost-Efficient Approach for Improving Multilingual CLIP Performance on Low-Resource Languages~\citep{dos_santos_capivara_2023}
    \item GemBode and PhiBode: Adapting Small Language Models to Brazilian Portuguese~\citep{garcia_gembode_2025}
    \item Sabiá: Portuguese Large Language Models~\citep{pires_sabia_2023}
\end{enumerate}

Based on this calibration, the search string used in this study was \textbf{(``language model'' OR ((``vision-language'' OR ``audio-language'' OR ``multimodal'') AND ``model'')) AND ``portuguese''}.

Exclusively on arXiv, we performed the automated search by breaking the search string into four parts (\textit{i.e.}, we performed four separate searches), as the platform is limited regarding the use of more complex expressions in search strings:
\begin{itemize}
    \item ``language model'' AND ``portuguese''
    \item ``audio-language model'' AND ``portuguese''
    \item ``vision-language model'' AND ``portuguese''
    \item ``multimodal model'' AND ``portuguese''
\end{itemize}

In all libraries, we used the year restriction because this study only considers papers published since 2020. We chose not to perform additional searches with the string in Portuguese because (i) currently, most papers on language models are published in English, and (ii) when published in Portuguese, the paper usually has an abstract in English\footnote{For example, the Brazilian Computer Society (SBC, from Portuguese: Sociedade Brasileira de Computação) requires an abstract in English for the publication of papers in Portuguese.}.

For all papers included by the previous step, we performed the snowballing process, which consisted of the following two steps: i) \textit{Forward}: Apply the inclusion and exclusion criteria, based on the title, to all papers that cite the included papers; and ii) \textit{Backward}: Apply the inclusion and exclusion criteria, based on the title, to all papers that were cited by the included papers.

\subsection{Inclusion and Exclusion Criteria}
\label{sec:IE}
We defined the inclusion (I) and exclusion (E) criteria as:

\begin{enumerate}[{\hspace{0.3cm} I}1.]
    \item Papers that propose a new language model, with one or more modalities, focusing on the Portuguese language.
    \item The proposed model must have been trained (pre-trained or fine-tuned) on textual data in Portuguese.
\end{enumerate}
\begin{enumerate}[{\hspace{0.3cm} E}1.]
    \item Papers not written in English or Portuguese.
    \item Papers that only propose datasets or benchmarks for language models.
    \item Papers that only propose approaches, techniques, and/or methods using or not using existing language models to solve a task.
    \item Papers that only describe the use, application, or evaluation of one or more existing language models or techniques (not language models).
    \item When the same model was reported more than once, only the most cited version will be kept.
    \item Sources that are not a paper, such as conference or workshop prefaces.
\end{enumerate}

\subsection{Selection Procedure}
Two reviewers (Reviewer 1 and Reviewer 2) screened all papers returned by the automated search against the inclusion and exclusion criteria. Following the criteria, each reviewer categorized the papers as Include, Exclude, or Uncertain. The selection procedure was guided by Table~\ref{tab:1}. Papers classified as A were included in the review, while those classified as B, C, and D were read in full to determine their eligibility for inclusion or exclusion. Papers classified as E were excluded from the study. We followed the same process during the snowballing stage.

\begin{table}[ht]
    \centering
    \small
    \renewcommand{\arraystretch}{1.2}
    \caption{Selection process for papers returned in the mapping searches.}
    \begin{tabular}{ccccc} \midrule
        \multicolumn{2}{c}{} & \multicolumn{3}{c}{\textbf{Reviewer 1}}\\\cmidrule{3-5}
        \multicolumn{2}{c}{} & Include & Uncertain & Exclude \\\midrule
        \multirow{4}{*}{\textbf{Reviewer 2}} & Include & \cellcolor{OliveGreen!100}\color{white} A & \cellcolor{Gray!25} B & \cellcolor{Gray!25} C\\\cmidrule{2-5}
        & Uncertain & \cellcolor{Gray!25} B & \cellcolor{Gray!25} C & \cellcolor{Gray!25} D \\\cmidrule{2-5}
        & Exclude & \cellcolor{Gray!25} C  & \cellcolor{Gray!25} D & \cellcolor{BrickRed!100}\color{white} E\\\midrule
    \end{tabular}
    \label{tab:1}
\end{table}

\subsection{Information Extraction}
\label{sec:exd}
Table~\ref{tab:2} shows the information extraction form developed to answer the research questions of this study. We extracted this information from the documentation of the identified models, which may include the paper or technical report introducing the model, or the model report in Model Cards or README files in repositories such as HuggingFace or GitHub. The following sections refer to this documentation as `study' or `paper'. 
Two reviewers reviewed all the extracted information. 

\begin{table}[!ht]
    \centering
    \renewcommand{\arraystretch}{1.6}
    \small
    \caption{Information Extraction Form from language models for Portuguese included in the mapping.}
    \begin{tabular}{>{\raggedright\arraybackslash}p{0.20\textwidth}>{\raggedright\arraybackslash}p{0.52\textwidth}>{\raggedright\arraybackslash}p{0.12\textwidth}}\hline
        \textbf{Field} & \textbf{Description} & \textbf{Related Q}\\\hline

        Bibliographic information & Title, Authors, Publication date, Publication venue, License, Abstract & N/A \\

        Contextual information & Authors' affiliation, Country, Type of publication (\textit{paper or technical report}), Code available? (\textit{yes or no}), Weights available? (\textit{yes or no}), Data available? (\textit{yes or no)} & Q2 \\

        Information about language models for Portuguese & Model name, Number of parameters, Base model name (if any), Target task(s), Portuguese variant (training) & Q1 \\

        Information about training data & Name of Benchmark(s) and/or Dataset(s), Originated in Portuguese and/or translated and/or generated data? & Q3 and Q4 \\\hline        
    \end{tabular}
    \label{tab:2}
\end{table}

\subsection{Quality Appraisal Criteria}
We evaluated each language model included in this mapping based on the quality of its documentation for reporting information about the proposed model. Table~\ref{tab:25} presents the quality assessment criteria used. We divided the quality criteria into three categories: Questions about the study; Questions about the language model development; and, Questions about the analysis of the language model. For each criterion, we assigned a score of `0'  if the information was not provided, `0.5' if it was partially provided, and `1' if it was completely provided. The total score that a paper can achieve in this study is 11.

\begin{table}[!ht]
    \centering
    \renewcommand{\arraystretch}{1.4}
    \small
    \caption{Quality Appraisal Criteria for language models included in the mapping.}
    \begin{tabular}{>{\raggedright\arraybackslash}p{0.8\textwidth}} \hline
        \textbf{Quality Criteria} \\\hline
        \textit{Category: Questions about the study}\\
        \textbf{QC1:} Do the authors clearly describe the objective(s) of the study? \\
        \textbf{QC2:} Do the authors clearly describe the study design?\\
        \textbf{QC3:} Do the authors clearly describe the limitation(s) of the study?\\
        \textbf{QC4:} Do the authors clearly describe the ethical considerations of the study? \\
        \hline
        \textit{Category: Questions about language model development}\\
        \textbf{QC5:} Do the authors describe the datasets used?\\
        \textbf{QC6:} Do the authors describe the data preprocessing procedures?\\
        \textbf{QC7:} Do the authors describe the model architecture? \\
        \hline
        \textit{Category: Questions about language model analysis}\\
        \textbf{QC8:} Do the authors describe the evaluation metrics used?\\
        \textbf{QC9:} Do the authors perform a quantitative analysis of the results? \\
        \textbf{QC10:} Do the authors compare the results with other existing models?\\
        \textbf{QC11:} Do the authors perform a qualitative analysis of the results?\\
        \hline
        \textbf{Score for each criterion:} No = 0, Partially = 0.5, Yes = 1
        \\\hline
    \end{tabular}
    \label{tab:25} 
\end{table}
\section{Review and Findings}
\label{sec:findings}

This section presents the execution and results of the systematic mapping study, providing an overview of the 46~included language models for Portuguese. The analysis of the extracted information, along with discussion, will be presented in Section~\ref{sec:ad}.

\subsection{Conducting the Review}

\subsubsection{Automatic Search Process}
We performed the search procedure on June 3, 2025, in all the libraries mentioned. Figure~\ref{fig:searchProcess} shows in the Automatic Search section the number of papers resulting in each step, from the ones retrieved in each source until the number of final papers screened by the inclusion and exclusion criteria after removing duplicates and entries that are not papers (such as conference reviews). In total, 32 language models for Portuguese were included in this stage, and all control papers were returned. 

\subsubsection{Snowballing}
\begin{figure}[!htb]
    \centering
    \includegraphics[width=0.95\linewidth, trim={0cm, 15.5cm, 0cm, 0cm}, clip]{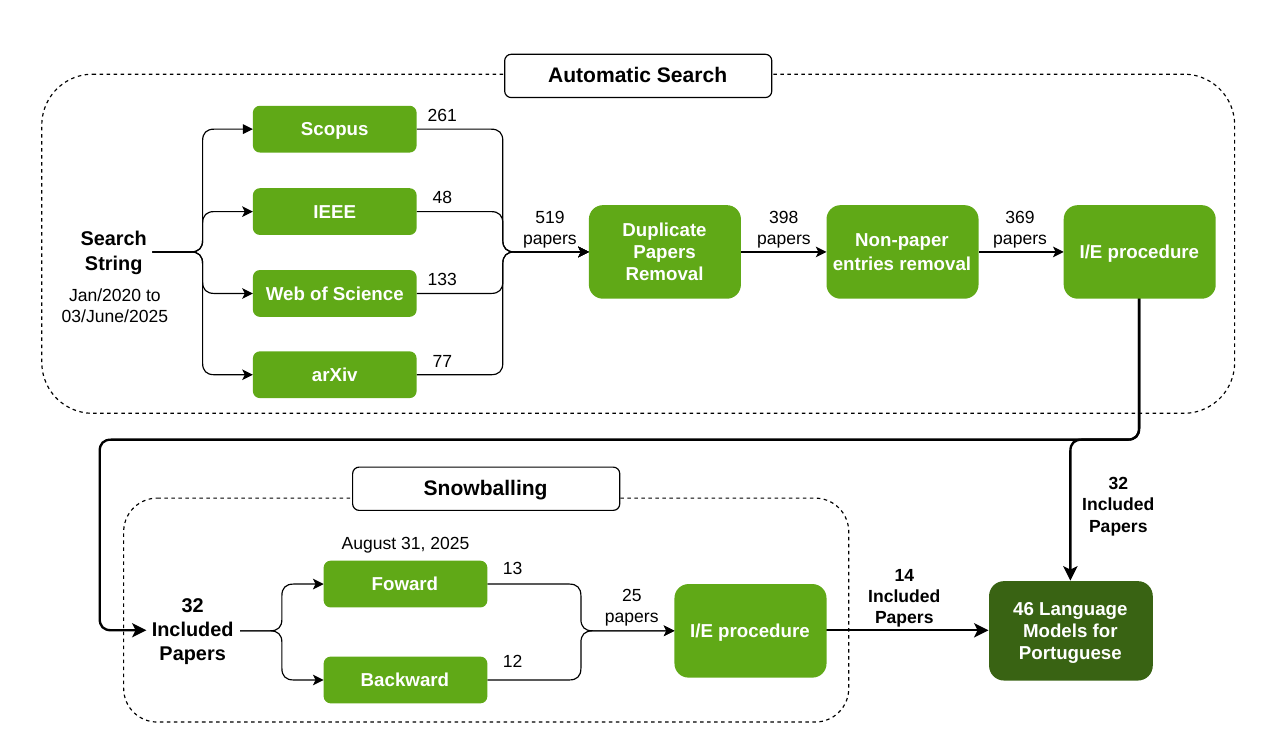}
    \caption{Papers identified by automatic search process and snowballing search.}
    \label{fig:searchProcess}
\end{figure}

We used the 32 papers identified in the automated search to conduct a (forward and backward) snowballing process. The procedure for collecting papers and works that cited the 32 papers was carried out on August~31, 2025. For each paper, two reviewers applied the selection criteria using the titles of the reference papers (backward) and the citation papers (forward). Figure~\ref{fig:searchProcess} shows in the Snowballing block the number of papers selected in the forward and backward snowballing process, excluding references included in the automated search stage, and the number of new papers included after reading. In this stage, 14  language models for Portuguese were included, bringing the total to 46 models in this study.

\subsection{Overview}
\label{sec:over}

\begin{figure}[!ht]
    \centering
    \includegraphics[width=0.98\linewidth, trim={0.3cm, 0.5cm, 0.3cm, 0.3cm}, clip]{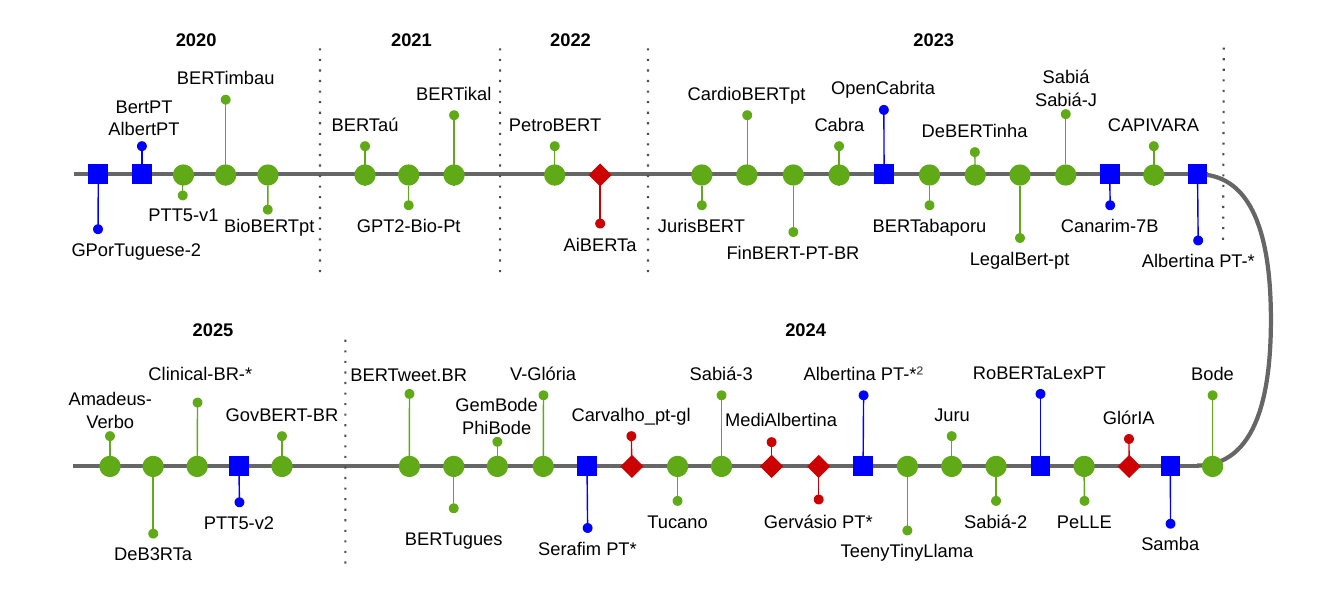}
    \caption{Language models for Portuguese released from 2020 to August 2025. \includegraphics[width=0.28cm, trim={0 0.3cm 0.6cm 0}, clip]{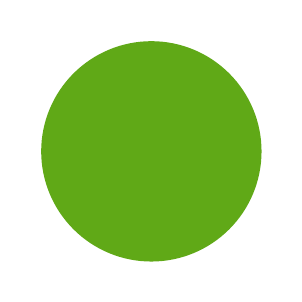}~Brazilian Portuguese, \includegraphics[width=0.3cm, trim={0cm 0.3cmcm 0.8cm 0}, clip]{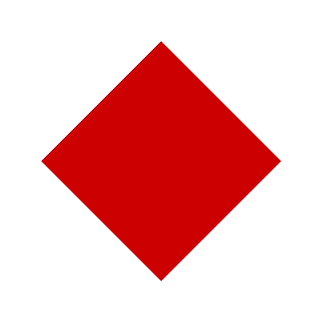}~European Portuguese, \includegraphics[width=0.28cm, trim={0 0.3cm 0.8cm 0}, clip]{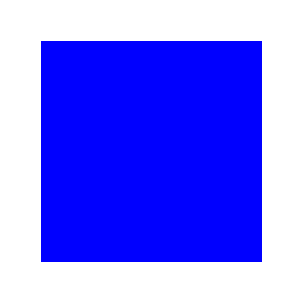}~both/general Portuguese variants.}
    \label{fig:lmsano}
\end{figure}

The publication period of the included works ranges from 2020\footnote{Selection criteria.} to August 2025. Figure~\ref{fig:lmsano} shows the language models by year of publication. The figure also indicates whether the model was proposed for Brazilian Portuguese, European Portuguese, or both/general Portuguese variants. Most models were published from 2023 onward. 

The publication venues, listed in Table~\ref{tab:lfp}, are diverse, with almost all related to the theme of this study. The BRACIS, EPIA, and PROPOR conferences stand out for having the highest numbers of publications: 6, 3, and 3, respectively. SIGUL, a special group on Under-resourced Languages, and CBMS, a symposium on computational medical systems, appeared with two publications. The latter is noteworthy because, while most venues are about AI or topics related to Natural Language Processing (NLP), there are a total of 4 publications~\citep{de_souza_pinto_developing_2025, schneider_gpt-2_2021, schneider_cardiobertpt_2023, schneider_biobertpt_2020} in CBMS, ClinicalNLP, and the \textit{Computers in Biology and Medicine} journal, which are venues exclusively for publications related to medicine. 

\begin{figure}[!ht]
    \centering
    \includegraphics[width=0.98\linewidth, trim={0cm, 2.6cm, 0cm, 1cm}, clip]{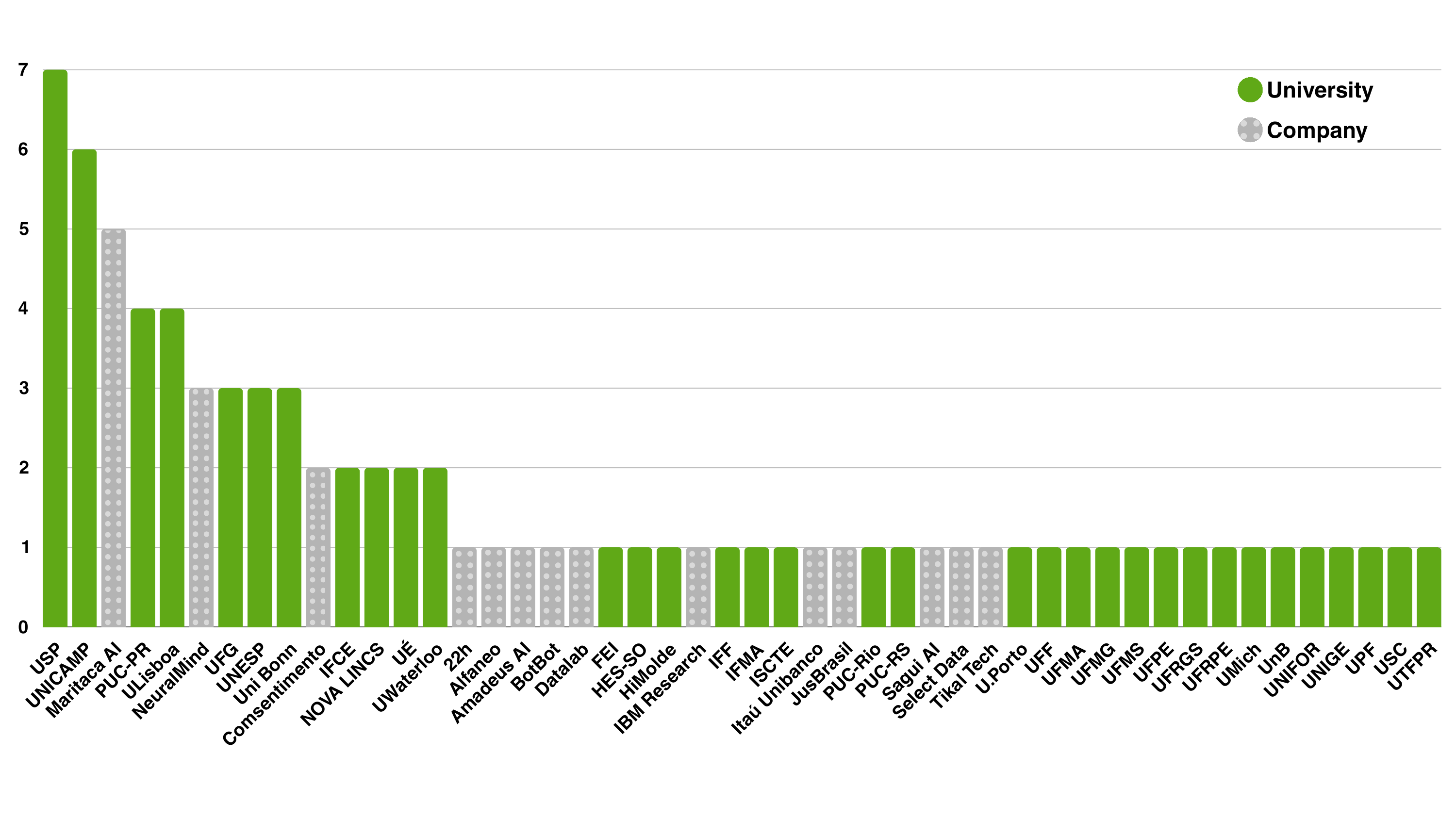}
    \caption{Authors' affiliations of language models for Portuguese. The acronyms can be found at Section~\ref{sec:acro}.}
    \label{fig:developers}
\end{figure}

\begin{figure}[!ht]
    \centering
    \includegraphics[width=0.90\linewidth, trim={3cm, 8.5cm, 3cm, 7.8cm}, clip]{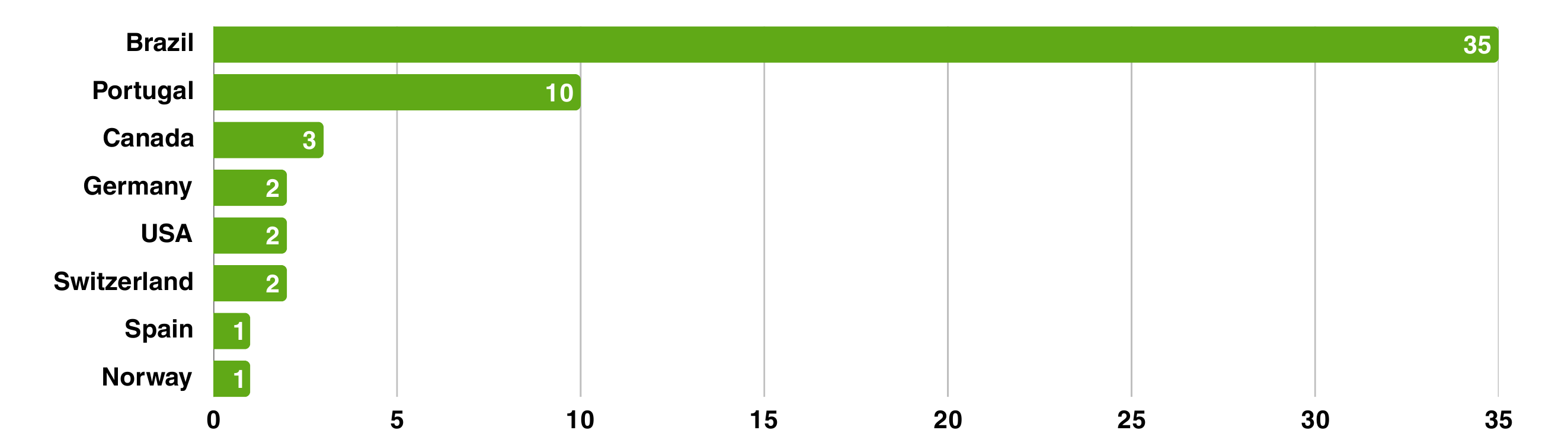}
    \caption{Institutions' country of language models for Portuguese.}
    \label{fig:country}
\end{figure}

The affiliations of the model developers are also diverse, including both public and private universities (a total of~33) and companies (a total of 14)\footnote{Note that more than one institution may have collaborated on the development of a single language model.}. Figure~\ref{fig:developers} shows the affiliations of the developers who proposed the models and the number of models developed in each institution. The institutions with the highest number of developed language models were the Brazilian public universities USP and UNICAMP, with 7 and 6 models, respectively, followed by the Brazilian company Maritaca AI with 5 models. Additionally, Figure~\ref{fig:country} shows the list of countries of the developers' institutions and the number of models developed in each country\footnote{Note that more than one country may have collaborated on the development of a single language model.}. The countries with the highest numbers of developed language models were Brazil, with 35 models, and Portugal, with 10 models. Also appearing on the list are Canada with 3 models, Germany, USA, and Switzerland, each with 2 models, and Spain and Norway, each with 1 model.

\begin{table}[!hbt]
    \centering
    \renewcommand{\arraystretch}{1.3}
    \small
    \caption{Publications venues and frequency for the identified language models.}
    \begin{tabular}{>{\raggedright\arraybackslash}p{0.835\textwidth}>{\centering\arraybackslash}p{0.1\textwidth}} \hline
        \textbf{Publication venue} & \textbf{Frequency} \\\hline

        \multicolumn{2}{c}{\cellcolor{Gray!10}\textbf{Conference papers}}  \\
        
        Brazilian Conference on Intelligent Systems (BRACIS)  & 6 \\
        
        Conference on Artificial Intelligence (EPIA) & 3 \\
        
        International Conference on Computational Processing of Portuguese Language (PROPOR) & 3 \\
        
        International Symposium on Computer-Based Medical Systems (CBMS) & 2 \\

        Special Interest Group on Under-resourced Languages (SIGUL) & 2 \\
        
        Brazilian Workshop on Artificial Intelligence in Finance (BWAIF) & 1 \\

        Clinical Natural Language Processing Workshop (ClinicalNLP) & 1 \\

        Computational Science and Its Applications (ICCSA)  & 1 \\
        
        Encontro Nacional de Inteligência Artificial and Computacional (ENIAC) & 1 \\
        
        Iberoamerican Congress on Pattern Recognition (CIARP) & 1 \\
        
        International Conference on Intelligent Data Engineering and Automated Learning (IDEAL) & 1 \\
        
        International Conference on Recent Advances in Natural Language Processing (RANLP) & 1 \\
        
        Workshop on Customizable NLP: Progress and Challenges in Customizing NLP for a Domain, Application, Group, or Individual (CustomNLP4U) & 1 \\
        
        Workshop on Multi-lingual Representation Learning (MRL) & 1 \\
        \arrayrulecolor{black!35} \hline
        
        \textbf{Total of conference papers  }& \textbf{25} \\\hline
        
        \multicolumn{2}{c}{\cellcolor{Gray!10}\textbf{Journal papers}}  \\
        
        Big Data and Cognitive Computing & 1 \\
        
        Computers in Biology and Medicine & 1 \\
        
        Machine Learning with Applications & 1 \\
        
        Neural Computing and Applications & 1 \\
        
        Semina: Ciências Exatas and Tecnológicas & 1 \\
        
        \arrayrulecolor{black!35} \hline
        
        \textbf{Total of journal papers  }& \textbf{5} \\\hline

        \multicolumn{2}{c}{\cellcolor{Gray!10}\textbf{Not peer-reviewed models}}  \\
        
        arXiv & 12\\
        
        HuggingFace & 4\\\hline
        
        \textbf{Total of not peer-reviewed models}& \textbf{16} \\
        
        \arrayrulecolor{black} \hline
        \cellcolor{Gray!25}\textbf{Total} & \cellcolor{Gray!25}\textbf{46} \\\hline
    \end{tabular}
    \label{tab:lfp}
\end{table}

Table~\ref{tab:mi2} presents the language models for Portuguese identified in this study. It characterizes each of the 46 models mapped by year of publication/release\footnote{We consider the publication date of the paper or the first version on arXiv or HuggingFace. In some cases, the model was released prior to the reported publication date.}, base model, purpose, number of parameters, license for use, and dataset used in the training stage.

{\footnotesize
\renewcommand{\arraystretch}{1.7}
\arrayrulecolor{black!5}

\begin{longtable}{>{\raggedright\arraybackslash}p{0.14\textwidth}|>
    {\raggedright\arraybackslash}p{0.11\textwidth}|>
    {\raggedright\arraybackslash}p{0.11\textwidth}|>
    {\raggedright\arraybackslash}p{0.11\textwidth}|>
    {\raggedright\arraybackslash}p{0.08\textwidth}|>
    {\raggedright\arraybackslash}p{0.26\textwidth}}
    \caption{Identified Language Models for Portuguese from 2020 to 2025.}
    \label{tab:mi2} \\
    \arrayrulecolor{black} \hline
    
    \textbf{Language Model} & \textbf{Base Model} & \textbf{Purpose} & \textbf{\#~Parameters} & \textbf{License} & \textbf{Training Data} \\\hline \arrayrulecolor{black!5}

      GPorTuguese-2 \citep{pierre2020gpt2smallportuguese} & GPT-2 \citep{radforda} & General & 124M & MIT License & Wikipedia-PT \citep{wikidump} \\\hline
        
      BertPT, AlbertPT \citep{feijo_mono_2020} & BERT \citep{devlin2019bert}, ALBERT \citep{lan2020albertlitebertselfsupervised} & General & 110M, 12M & GNU General Public License v3.0  & Wikipedia-PT \citep{wikidump}, Open Subtitles \citep{lison-tiedemann-2016-opensubtitles2016}, CHAVE \citep{CHAVE_dataset}, Folha de São Paulo \citep{marlesson_folhauol_news}, EuroParl \citep{koehn2005europarl}, Research abstracts \citep{dominio_publico_portal}
      \\\hline
        
      PTT5-v1 \citep{carmo_ptt5_2020} & T5 \citep{t5} & General & 60M, 220M, 740M & MIT License & brWaC \citep{wagner2018brwac} \\\hline
        
      BERTimbau \citep{souza_bertimbau_2020} & BERT \citep{devlin2019bert} & General & 110M, 335M  & MIT License & brWaC \citep{wagner2018brwac} \\\hline
        
      BioBERTpt \citep{schneider_biobertpt_2020} & mBERT \citep{devlin2019bert} & General in clinical domain & Not reported & Not reported  & EHR from Brazilian hospitals \citep{ehr_brazil}, Literature titles and abstracts from Scielo \citep{scielo_dataset} and Pubmed \citep{pubmed_dataset}   
      \\\hline
        
      BERTaú \citep{bertau} & BERT \citep{devlin2019bert} & General in financial domain & Not reported & Not reported & Proprietary conversational data (AVI – Assistente Virtual Itaú) \citep{bertau}
      \\\hline

      GPT2-Bio-Pt \citep{schneider_gpt-2_2021} & GPT-2 \citep{radforda} & General in clinical and biomedical domain & Not reported & Not reported  & Literature titles and abstracts from Scielo \citep{scielo_dataset} and Pubmed \citep{pubmed_dataset}
      \\\hline
        
      BERTikal \citep{polo_legalnlp_2021} & BERTimbau \citep{souza_bertimbau_2020} & General in legal domain & 110M, 335M  & MIT License  & Brazilian legal documents (court publications, case movements, and TJSP records) \citep{polo_legalnlp_2021}
      \\\hline
        
      PetroBERT \citep{petrobert} & mBERT \citep{devlin2019bert}, BERTimbau \citep{souza_bertimbau_2020} & General in oil and gas domain & Not reported & Not reported  & GeoCorpus \citep{amaral2017geocorpus}, Petrolês \citep{gomes2021petroles} and DDR-Corpus \citep{ddr_corpus}
      \\\hline
      
      AiBERTa \citep{miquelina_generating_2022} & BERT \citep{devlin2019bert} & General & Not reported, but see HuggingFace & MIT License & European Portuguese corpus from Arquivo.pt \citep{miquelina_generating_2022, arquivo_pt}
      \\\hline
      
      JurisBERT \citep{viegas_jurisbert_2023} & BERT \citep{devlin2019bert} &General in legal domain & 110M & The Unlicense & Custom Brazilian legal corpus (laws, court decisions, and legal texts) \citep{viegas_jurisbert_2023}
      \\\hline
      
      CardioBERTpt \citep{schneider_cardiobertpt_2023} & BERTimbau \citep{souza_bertimbau_2020}, mBERT \citep{devlin2019bert}, BioBERTpt \citep{schneider_biobertpt_2020}  & General in cardiology domain & Not reported & Apache license 2.0  & TempClinBr \citep{tempclinbr}  SemClinBr \citep{semclinbr}, Private cardiology clinical corpus \citep{schneider_cardiobertpt_2023} 
      \\\hline
      
      FinBERT-PT-BR \citep{santos_finbert-pt-br_2023} & BERTimbau \citep{souza_bertimbau_2020} & General in financial domain& Not reported& Apache license 2.0   & Brazilian financial news corpus (Valor Econômico, Exame, InfoMoney) \citep{santos_finbert-pt-br_2023}
      \\\hline

      Cabra \citep{botbotroboticscabra_2024} & Llama 2 \citep{llama2} & General & 7B & Not reported  & PortugueseDolly \citep{portuguesedolly}
      \\\hline
      
      OpenCabrita \citep{larcher_cabrita_2023} & OpenLlama \citep{openlm2023openllama} & General & 3B &Apache 2.0 license & mC4 \citep{xue-etal-2021-mt5}
      \\\hline
      
      BERTabaporu \citep{da_costa_bertabaporu_2023} & BERT \citep{devlin2019bert} & General in tweets domain & 110M, 335M & MIT License & Brazilian Twitter corpus \citep{da_costa_bertabaporu_2023}
      \\\hline
      
      DeBERTinha \citep{debertinha} & DeBERTaV3 \citep{he2023debertav3improvingdebertausing} & General & 40M & MIT License  & brWaC \citep{wagner2018brwac}, Carolina \citep{crespo2023carolinageneralcorpuscontemporary}
      \\\hline
      
      LegalBert-pt \citep{silveira_legalbert} & BERT \citep{devlin2019bert}, BERTimbau \citep{souza_bertimbau_2020} &  General in legal domain & 110M & Responsible AI License & Brazilian legal corpus (from 10 courts via CNJ Codex system) \citep{silveira_legalbert}
      \\\hline
      
      Sabiá, Sabiá-J \citep{pires_sabia_2023} & Llama \citep{llama1}, GPT-J \citep{wang2021gpt}  & General & Not reported & same as Llama-1  &  ClueWeb22 \citep{Overwijk:ClueWeb22} (Portuguese subset, filtered) \\\hline
      
      Canarim-7B \citep{maicon_domingues_2023} & Llama 2 \citep{llama2} & General & 7B & Llama2  & CommonCrawl \citep{commoncrawl}(CC-MAIN-2023-23, Portuguese subset)  \\\hline
      
      CAPIVARA \citep{dos_santos_capivara_2023} & OpenCLIP \citep{ilharco_2021_5143773} & General in vision language tasks & 1.9M, 278M & MIT License &  CC3M \citep{Sharma2018ConceptualCA}
      \\\hline
      
      Albertina PT-* \citep{albertina} & DeBERTa \citep{he2020deberta} & General & 900M & MIT License &  brWaC \citep{wagner2018brwac}, OSCAR \citep{abadji-etal-2022-towards}, DCEP \citep{hajlaoui-etal-2014-dcep}, Europarl \citep{koehn2005europarl}, ParlamentoPT (Portuguese parliamentary debates corpus) \citep{albertina}
      \\\hline
    
      Bode \citep{garcia_introducing_2024} & Llama 2 \citep{llama2} &General &7B and 13B &MIT& Portuguese-translated version of Alpaca~ \cite{alpaca_dataset}
      \\\hline

      Samba \citep{samba} & Llama 2 \citep{llama2} & General & 1.1B & Academic Free License v3.0  & Not reported
      \\\hline
      
      GlórIA \citep{lopes2024gloria} & GPTNeo \citep{black2021gpt} &General &1.3B and 2.7B & restricted to research-only purposes & OSCAR \citep{abadji-etal-2022-towards}, ClueWeb-L~22 \citep{Overwijk:ClueWeb22}, Open Subtitles \citep{lison-tiedemann-2016-opensubtitles2016}, Wikipedia-PT \citep{wikidump}, Europarl \citep{koehn2005europarl}and Arquivo.pt \citep{arquivo_pt}
      \\\hline
      
      PeLLE \citep{demello2024pelle} & RoBERTa \citep{liu2019robertarobustlyoptimizedbert}, XML-R \citep{conneau-etal-2020-unsupervised}, mBERT \citep{devlin2019bert} &  General &Not reported &Not reported & Carolina \citep{crespo2023carolinageneralcorpuscontemporary}
      \\\hline

      RoBERTaLexPT \citep{garcia_robertalexpt_2024} & RoBERTa \citep{liu2019robertarobustlyoptimizedbert} & General in legal domain & 125M, 355M  &Creative Commons Corporation CC-BY-4.0 & LegalPT corpus (composed of MultiLegalPile \citep{niklaus2024multilegalpile689gbmultilinguallegal}, Ulysses-Tesemõ \citep{Tesemo2024}, ParlamentoPT \citep{albertina}, Iudicium Textum \citep{IudiciumTextum}, Acordãos TCU \citep{Bonifacio2020ASO}, DataSTF \citep{datastf}), CrawlPT corpus (composed of brWaC \citep{wagner2018brwac}, CC100 \citep{conneau-etal-2020-unsupervised}, OSCAR \citep{abadji-etal-2022-towards})
      \\\hline
      
      Sabiá-2 \citep{almeida_sabia-2_2024} & Not reported &  General &Not reported &Not reported & Not reported
      \\\hline
      
      Juru \citep{junior2024juru} & Mistral \citep{jiang2023mistral} &General in legal domain &7B &Not reported & LexML \citep{lexml_brasil}, Dados~Abertos \citep{Sakiyama:DadosAbertos} and Portuguese Legal Academic Papers \citep{br_legal_academic_corpus} (Brazil, web-scraped)
      \\\hline
      
      TeenyTinyLlama \citep{teenytinyllama} & Llama 2 \citep{llama2} &General &160M, 460M & Apache 2.0 license & Wikipedia-PT \citep{wikidump}, OSCAR \citep{abadji-etal-2022-towards}, ROOTS \citep{10.5555/3600270.3602576}, Common Crawl \citep{wenzek-etal-2020-ccnet}, CulturaX \citep{nguyen-etal-2024-culturax}, Portuguese-translated versions of Instruct-PTBR \citep{instruct_ptbr_enus_11m}, Gpt4all-J \citep{gpt4all_j_pt}, Bactrian-X \citep{BactrianX}, Dolly~15K\cite{portuguesedolly}, CosmosQA \citep{huang-etal-2019-cosmos}
      \\\hline

      Albertina PT-*$^2$ \citep{santos-etal-2024-fostering} & DeBERTa \citep{he2020deberta} & General & 100M, 1.5B & MIT License & OSCAR \citep{abadji-etal-2022-towards}, CulturaX \citep{nguyen-etal-2024-culturax}, DCEP \citep{hajlaoui-etal-2014-dcep}, Europarl \citep{koehn2005europarl}, ParlamentoPT \citep{albertina}
      \\\hline
      
      Gervásio PT* \citep{santos_advancing_2024} & Llama 2 \citep{llama2}  &General & 8B &MIT License & GLUE \citep{wang-etal-2018-glue}, SuperGLUE \citep{10.5555/3454287.3454581}
      \\\hline
      
      MediAlbertina \citep{nunes_medialbertina_2024} & Albertina PT-PT \citep{albertina}  & General in medical domain &900M &MIT License  & Private clinical EMR corpus (Hospital de Santa Maria) \citep{nunes_medialbertina_2024}
      \\\hline
      
      Sabiá-3 \citep{abonizio_sabia-3_2025} & Not reported &General & Not reported & Not reported & Not reported
      \\\hline
      
      Tucano \citep{correa_tucano_2024} & Llama \citep{llama1} &General & 160M, 630M, 1.1B, and 2.4B & Apache 2.0 license & GigaVerbo \citep{correa_tucano_2024} (several portions of openly available datasets for Portuguese)
      \\\hline
      
      Carvalho\_pt-gl \citep{gamallo_galician-portuguese_2025} & Cerebras-GPT \citep{dey2023cerebrasgptopencomputeoptimallanguage} &General &1.3B &MIT License & CorpusNÓS \citep{de-dios-flores-etal-2024-corpusnos}, BNE web corpus (Galician subset) \citep{bne_web_corpus}, Arquivo.pt \citep{arquivo_pt}
      \\\hline
      
      Serafim PT* \citep{gomes_open_2025} & SBERT \citep{reimers2019sentence} &General & 100M, 335M, 900M &MIT License& EuroParl \citep{koehn2005europarl}, EUbookshop \citep{eu_bookshop}, TED~2020 \citep{reimers-gurevych-2020-making}, Tatoeba \citep{tatoeba}
      \\\hline
      
      V-GlórIA \citep{simplicio_v-gloria_2024} & FROMAGe model \citep{koh2023groundinglanguagemodelsimages}  &General in vision language tasks &Not reported& & CC3M \citep{Sharma2018ConceptualCA} \\\hline

      GemBode, PhiBode \citep{garcia_gembode_2025} & Gemma \citep{team2024gemma}, Phi \citep{li2023textbooks, abdin2024phi3technicalreporthighly} &General &PhiBode: 1.5B, 2B, 3B; GemBode: 2B, 7B &Apache license 2.0 & UltraAlpaca \citep{garcia_gembode_2025}
      \\\hline
      
      BERTugues \citep{zago_bertugues_2024} & BERTimbau \citep{souza_bertimbau_2020} & General &110M &Not reported & Wikipedia-PT \citep{wikidump}, brWaC \citep{wagner2018brwac} 
      \\\hline
      
      BERTweet.BR \citep{carneiro_bertweetbr_2025} & RoBERTa \citep{liu2019robertarobustlyoptimizedbert} &General in tweets domain &110 M  &Apache license 2.0 & Portuguese Twitter corpus (Internet Archive) \citep{twitter_archive_pt}
      \\\hline
      
      GovBERT-BR \citep{silva_govbert-br_2025} & BERTimbau \citep{souza_bertimbau_2020} &General in governmental data domain&Not reported &Not reported & Administrative~Data \citep{Administrative_Data_sbbd}, VICTOR \citep{luz-de-araujo-etal-2020-victor}
      \\\hline
      
      PTT5-v2 \citep{piau_ptt5-v2_2025} & T5 \citep{t5} &General &60M, 220M, 740M, 3B. &Apache license 2.0 & mC4 \citep{xue-etal-2021-mt5}
      \\\hline
      
      Clinical-BR-* \citep{de_souza_pinto_developing_2025} & Llama 2 \citep{llama2}, Mistral \citep{jiang2023mistral} &General in clinical notes domain  &7B &Apache license 2.0& SemClinBr \citep{semclinbr}, BRATECA \citep{PhysioNet-brateca-1.1}, PortugueseClinicalNER \citep{lopes-etal-2019-contributions}
      \\\hline
      
      DeB3RTa \citep{pires_deb3rta_2025} & DeBERTa \citep{he2020deberta} & General in financial domain &426M and 70M &MIT License& OFFCOMBR-3 \citep{OFFCOMBR-3}, FAKE.BR \citep{FAKE.BR}, CAROSIA \citep{CAROSIA}, BBRC \citep{BBRC}, Relevant~Fact \citep{cvm_financial_data}, Google~Patent \citep{google_patents}, Scielo \citep{scielo_dataset}, Wikipedia-PT \citep{wikidump}
      \\\hline
      
      Amadeus-Verbo \citep{cruz-castaneda_amadeus-verbo_2025} & Qwen2.5 \citep{qwen25} &  General &0.5B, 1.5B, 3B, 7B, 14B, 32B, 72B &Apache license 2.0 & Proprietary Instruction Dataset \citep{cruz-castaneda_amadeus-verbo_2025}(~600K)
      \\\arrayrulecolor{black}\hline

\end{longtable}
}
\section{Analysis and Discussion}
\label{sec:ad}

This section presents the analysis and discussion of the mapped language models developed for Portuguese. Section~\ref{sec:avql} presents the results of the quality assessment of the analyzed works. Section~\ref{sec:filo} presents the purposes and a phylogeny of the identified models.  Section~\ref{sec:disponibilidade} presents the availability of the mapped models, in terms of code, model, data, and documentation. Finally, Section~\ref{sec:cdavaliacao} presents an overview of the datasets and benchmarks used to train these models. 

\subsection{Quality Assessment}
\label{sec:avql}

\begin{table}[!ht]
    \caption{Quality assessment of included language models. QC -- Quality Criteria (Table~\ref{tab:25}). \mbox{P -- Total quality score.}}
    \centering
    \renewcommand{\arraystretch}{1.4}
    \scriptsize \arrayrulecolor{black}
    \setlength{\tabcolsep}{2.25pt} 
    \begin{tabular}{>{\raggedright\arraybackslash}p{0.29\textwidth}>{\centering\arraybackslash}p{0.046\textwidth}>{\centering\arraybackslash}p{0.046\textwidth}>{\centering\arraybackslash}p{0.046\textwidth}>{\centering\arraybackslash}p{0.046\textwidth}>{\centering\arraybackslash}p{0.046\textwidth}>{\centering\arraybackslash}p{0.046\textwidth}>{\centering\arraybackslash}p{0.046\textwidth}>{\centering\arraybackslash}p{0.046\textwidth}>{\centering\arraybackslash}p{0.046\textwidth}>{\centering\arraybackslash}p{0.046\textwidth}>{\centering\arraybackslash}p{0.046\textwidth}>{\centering\arraybackslash}p{0.046\textwidth}} \hline
     
    \textbf{Paper} & \textbf{QC1} & \textbf{QC2} & \textbf{QC3}& \textbf{QC4}& \textbf{QC5}& \textbf{QC6}& \textbf{QC7}& \textbf{QC8}& \textbf{QC9}& \textbf{QC10} & \textbf{QC11}& \textbf{P} \\\hline\arrayrulecolor{black!5}
    
        GPorTuguese-2\tiny~\citep{pierre2020gpt2smallportuguese} & 1 & 1 & 0.5 & 0 & 1 & 1 & 1 & 0.5 & 0.5 & 0.5 & 0.5 & 7.5 \\ \hline
        BertPT, AlbertPT\tiny~\citep{feijo_mono_2020} & 1 & 1 & 0 & 0 & 1 & 0.5 & 0.5 & 1 & 1 & 1 & 0 & 7 \\ \hline
        PTT5-v1\tiny~\citep{carmo_ptt5_2020} & 1 & 1 & 0 & 0 & 1 & 1 & 1 & 1 & 1 & 1 & 0 & 8 \\ \hline
        BERTimbau\tiny~\citep{souza_bertimbau_2020} & 1 & 1 & 0 & 0 & 1 & 0.5 & 1 & 1 & 1 & 1 & 0 & 7.5 \\ \hline
        BioBERTpt\tiny~\citep{schneider_biobertpt_2020} & 1 & 1 & 0 & 0 & 1 & 1 & 0.5 & 1 & 1 & 1 & 0 & 7.5 \\ \hline
        BERTaú\tiny~\citep{bertau} & 1 & 1 & 0 & 0 & 0.5 & 0.5 & 0.5 & 1 & 1 & 1 & 0 & 6.5 \\ \hline
        GPT2-Bio-Pt\tiny~\citep{schneider_gpt-2_2021} & 1 & 1 & 0.5 & 0 & 1 & 0.5 & 1 & 1 & 0.5 & 1 & 0.5 & 8 \\ \hline
        BERTikal\tiny~\citep{polo_legalnlp_2021} & 1 & 1 & 0 & 0 & 1 & 0.5 & 0.5 & 1 & 1 & 1 & 0 & 7 \\ \hline
        PetroBERT\tiny~\citep{petrobert}& 1 & 1 & 0 & 0 & 0.5 & 0 & 1 & 1 & 0.5 & 1 & 0 & 6 \\ \hline
        AiBERTa\tiny~\citep{miquelina_generating_2022} & 1 & 0.5 & 0 & 0 & 1 & 1 & 1 & 0 & 0 & 0 & 0 & 4.5 \\ \hline
        JurisBERT\tiny~\citep{viegas_jurisbert_2023} & 1 & 1 & 0 & 0 & 1 & 0.5 & 1 & 1 & 1 & 1 & 0 & 7.5 \\ \hline
        CardioBERTpt\tiny~\citep{schneider_cardiobertpt_2023} & 1 & 0.5 & 0 & 0 & 1 & 0 & 0.5 & 1 & 1 & 1 & 0 & 6 \\ \hline
        FinBERT-PT-BR\tiny~\citep{santos_finbert-pt-br_2023} & 1 & 1 & 0 & 0 & 1 & 0.5 & 0.5 & 1 & 0.5 & 0 & 0 & 5.5 \\ \hline
        Cabra 7b\tiny~\citep{botbotroboticscabra_2024} & 1 & 0 & 0 & 0 & 0.5 & 0 & 0.5 & 0 & 0 & 0 & 0 & 2 \\ \hline    
        OpenCabrita\tiny~\citep{larcher_cabrita_2023} & 1 & 1 & 0.5 & 0 & 1 & 0.5 & 1 & 1 & 1 & 1 & 0 & 8 \\ \hline
        BERTabaporu\tiny~\citep{da_costa_bertabaporu_2023} & 1 & 1 & 0 & 0 & 1 & 1 & 1 & 0.5 & 0.5 & 0.5 & 0 & 6.5 \\ \hline
        DeBERTinha\tiny~\citep{debertinha} & 1 & 1 & 0 & 0 & 1 & 0.5 & 0.5 & 1 & 1 & 1 & 0 & 7 \\ \hline
        LegalBert-pt\tiny~\citep{silveira_legalbert} & 1 & 1 & 0 & 0 & 1 & 0.5 & 1 & 1 & 1 & 1 & 0 & 7.5 \\ \hline
        Sabiá, Sabiá-J\tiny~\citep{pires_sabia_2023} & 1 & 1 & 1 & 0 & 1 & 0.5 & 1 & 1 & 1 & 1 & 0 & 8.5 \\ \hline
        Canarim-7B\tiny~\citep{maicon_domingues_2023} & 1 & 0.5 & 0 & 0 & 0.5 & 0 & 0.5 & 0.5 & 0.5 & 0 & 0 & 3.5 \\ \hline
        CAPIVARA\tiny~\citep{dos_santos_capivara_2023} & 1 & 1 & 1 & 1 & 1 & 1 & 1 & 1 & 1 & 0.5 & 1 & 10.5 \\ \hline
        Albertina PT-*\tiny~\citep{albertina} & 1 & 1 & 0 & 0 & 1 & 1 & 1 & 1 & 1 & 0.5 & 0 & 7.5 \\ \hline
        Bode\tiny~\citep{garcia_introducing_2024} & 1 & 1 & 0.5 & 0 & 0.5 & 0.5 & 1 & 1 & 1 & 1 & 0 & 7.5 \\ \hline
        Samba\tiny~\citep{samba} & 1 & 0 & 0 & 0 & 0 & 0 & 0.5 & 0 & 0 & 0 & 0 & 1.5 \\ \hline
        GlórIA\tiny~\citep{lopes2024gloria} & 1 & 1 & 1 & 0 & 1 & 1 & 1 & 1 & 1 & 1 & 1 & 10 \\ \hline
        PeLLE\tiny~\citep{demello2024pelle} & 1 & 1 & 0 & 0 & 1 & 0.5 & 0.5 & 1 & 1 & 1 & 0 & 7 \\ \hline
        RoBERTaLexPT\tiny~\citep{garcia_robertalexpt_2024} & 1 & 1 & 0 & 0 & 1 & 1 & 1 & 1 & 1 & 1 & 0 & 8 \\ \hline    
        Sabiá-2\tiny~\citep{almeida_sabia-2_2024} & 1 & 0 & 1 & 0 & 0.5 & 0 & 0 & 1 & 1 & 1 & 0 & 5.5 \\ \hline
        Juru\tiny~\citep{junior2024juru} & 1 & 1 & 1 & 0 & 1 & 0.5 & 1 & 1 & 1 & 0.5 & 0 & 8 \\ \hline
        TeenyTinyLlama\tiny~\citep{teenytinyllama} & 1 & 1 & 1 & 0 & 1 & 0.5 & 1 & 1 & 1 & 1 & 0 & 8.5 \\ \hline
        Albertina PT-*$^2$\tiny~\citep{santos-etal-2024-fostering} & 1 & 1 & 0 & 0 & 1 & 1 & 1 & 1 & 1 & 1 & 0 & 8 \\\hline
        Gervásio PT*\tiny~\citep{santos_advancing_2024} & 1 & 1 & 0.5 & 0.5 & 1 & 0.5 & 1 & 1 & 1 & 1 & 0 & 8.5 \\ \hline
        MediAlbertina\tiny~\citep{nunes_medialbertina_2024} & 1 & 1 & 0 & 0 & 1 & 0.5 & 1 & 1 & 1 & 1 & 0 & 7.5 \\ \hline
        Sabiá-3\tiny~\citep{abonizio_sabia-3_2025} & 1 & 0 & 0 & 0 & 0.5 & 0 & 0 & 1 & 1 & 1 & 0 & 4.5 \\ \hline
        Tucano\tiny~\citep{correa_tucano_2024} & 1 & 1 & 1 & 0.5 & 1 & 1 & 1 & 1 & 1 & 1 & 1 & 10.5 \\ \hline
        Carvalho\_pt-gl\tiny~\citep{gamallo_galician-portuguese_2025} & 1 & 1 & 0 & 0 & 1 & 1 & 1 & 1 & 1 & 1 & 0 & 8 \\ \hline
        Serafim PT*\tiny~\citep{gomes_open_2025} & 1 & 1 & 0 & 0 & 1 & 0.5 & 0.5 & 1 & 1 & 1 & 0 & 7 \\ \hline
        V-GlórIA\tiny~\citep{simplicio_v-gloria_2024} & 1 & 1 & 0.5 & 0.5 & 1 & 1 & 1 & 1 & 1 & 1 & 1 & 10 \\ \hline
        GemBode, PhiBode\tiny~\citep{garcia_gembode_2025} & 1 & 1 & 0.5 & 0 & 1 & 0.5 & 1 & 1 & 1 & 1 & 0 & 8 \\ \hline
        BERTugues\tiny~\citep{zago_bertugues_2024} & 1 & 1 & 0 & 0 & 1 & 0.5 & 0.5 & 1 & 1 & 1 & 0 & 7 \\ \hline
        BERTweet.BR\tiny~\citep{carneiro_bertweetbr_2025} & 1 & 1 & 1 & 0 & 1 & 1 & 1 & 1 & 1 & 1 & 1 & 10 \\ \hline
        GovBERT-BR\tiny~\citep{silva_govbert-br_2025} & 1 & 1 & 0 & 0 & 1 & 0.5 & 0.5 & 1 & 1 & 1 & 0 & 7 \\ \hline
        PTT5-v2\tiny~\citep{piau_ptt5-v2_2025} & 1 & 1 & 0 & 0 & 1 & 0.5 & 1 & 1 & 1 & 1 & 0 & 7.5 \\ \hline
        Clinical-BR-*\tiny~\citep{de_souza_pinto_developing_2025} & 1 & 1 & 0.5 & 0 & 1 & 0 & 1 & 1 & 1 & 1 & 0.5 & 8 \\ \hline
        DeB3RTa\tiny~\citep{pires_deb3rta_2025} & 1 & 1 & 1 & 0.5 & 1 & 0.5 & 1 & 1 & 1 & 1 & 0 & 9 \\ \hline
        Amadeus-Verbo\tiny~\citep{cruz-castaneda_amadeus-verbo_2025} & 0.5 & 0.5 & 0 & 0 & 0.5 & 0 & 0.5 & 0.5 & 1 & 0.5 & 0 & 4 \\ \hline\arrayrulecolor{black}
        \textbf{Average} & 0.99 & 0.87 & 0.28 & 0.07 & 0.89 & 0.55 & 0.79 & 0.89 & 0.87 & 0.83 & 0.14 & \textbf{7.5}\\ \hline
    \end{tabular}
    \label{tab:aqt}
\end{table}

The quality scores of the included language models, obtained using the quality criteria presented in Table~\ref{tab:25}, are reported in Table~\ref{tab:aqt}. Each study received a total score between 1.5 and 10.5, in 0.5-point intervals. The average score for all studies is 7.5 out of 11. These scores demonstrate the quality of these studies under different aspects and are not used to exclude any study. Among the 11~quality assessment criteria, those with the lowest average scores were, respectively, QC4 (\textit{Do the authors clearly describe the ethical considerations of the study?}) with 0.07, QC11 (\textit{Do the authors perform a qualitative analysis of the results?}) with 0.14, and QC3 (\textit{Do the authors clearly describe the limitation(s) of the study?}) with 0.28. These results show that, in the context of studies introducing language models for Portuguese, it is still not natural for researchers and developers to critically examine the ethical considerations and limitations of their model, as discussed in~\citep{jhessicaspringer, laai-ethics}. In addition, the studies usually do not provide a qualitative analysis of the results, which is very important for forming the basis of the critical discussion of criteria QC3 and QC4.

\subsection{Phylogeny and Purposes of Identified Models}
\label{sec:filo}

\begin{figure}[!tbh]
    \centering
    \includegraphics[width=0.99\linewidth, trim={0cm, 7cm, 0cm, 6cm},clip]{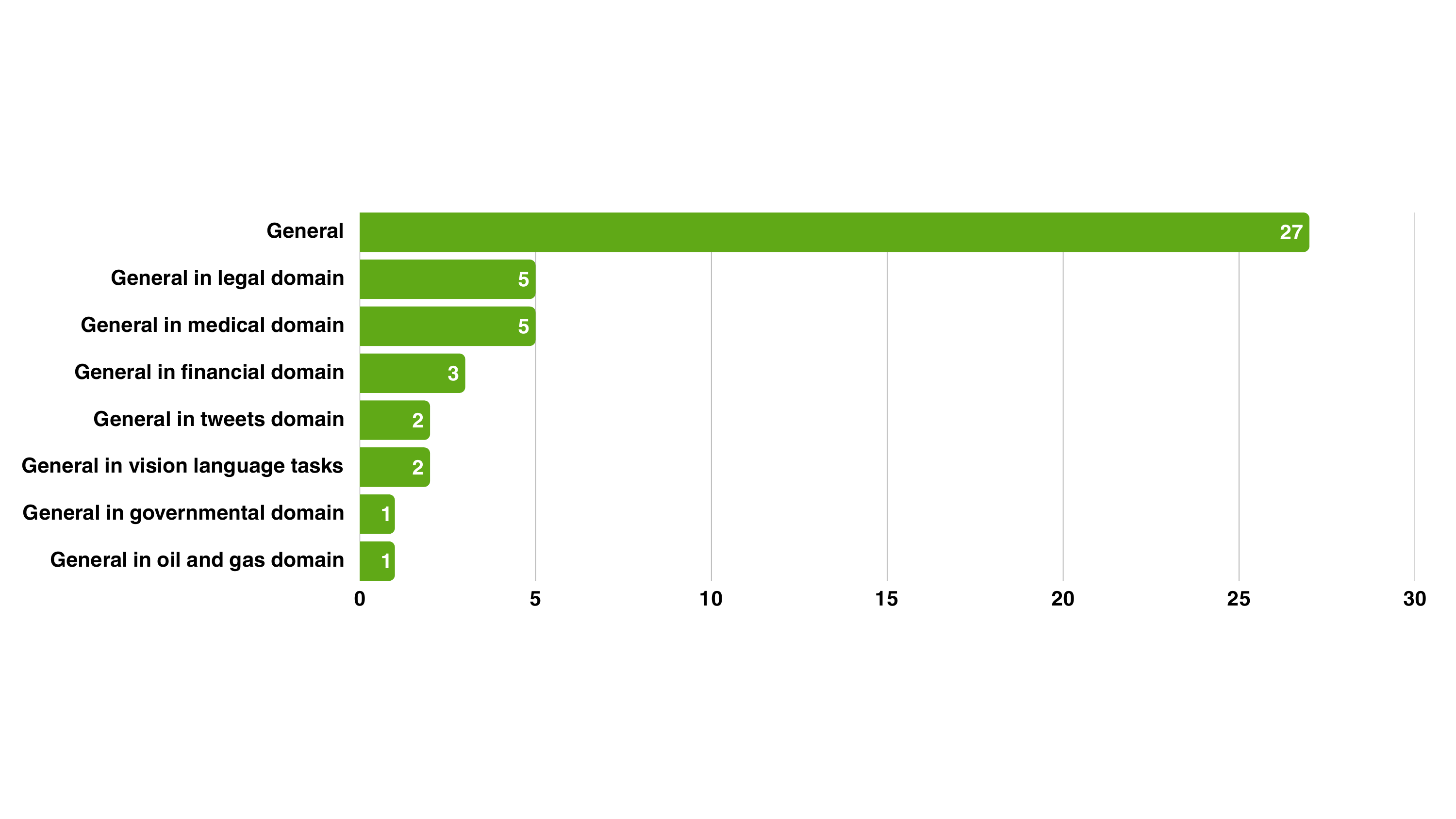}
    \caption{Overview of the purpose of language models for Portuguese. Here, the 'General in cardiology domain', 'General in clinical and biomedical domain', 'General in clinical domain', 'General in clinical notes domain', and 'General in medical domain' categories shown in the Table~\ref{tab:mi2},  have been merged into the 'General in medical domain' category.}
    \label{fig:purposes}
\end{figure}

\begin{figure}[p]
    \centering
    \includegraphics[width=0.97\linewidth, trim={0, 0.7cm, 3cm, 0}]{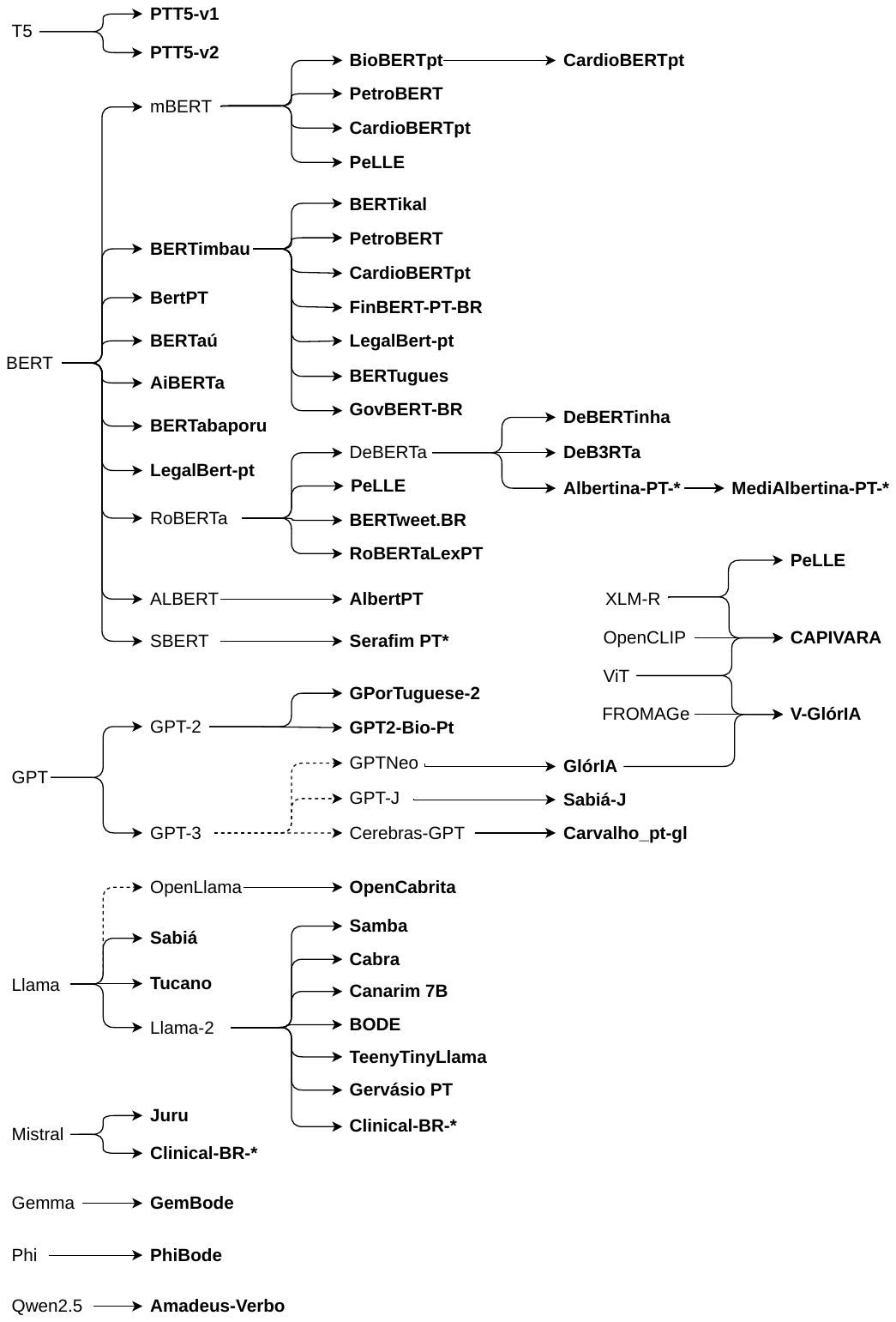}
    \caption{Phylogeny of language models for Portuguese (in bold). Straight lines represent models based on another. Dashed lines mean open models that reproduce another.}
    \label{fig:filo}
\end{figure}

The language models identified and presented in Section~\ref{sec:over} are diverse, differing in their purposes and approaches, although they all focus on Portuguese. Figure~\ref{fig:purposes} provides an overview of the purpose of the identified models (detailed specification for each model is presented in Table~\ref{tab:mi2}). Of the 46 identified models, 27 were trained for general tasks involving Portuguese-language text, and 2 focused on tasks that include both Portuguese-language text and images. The remaining models, although task-general, specialize in a specific domain. We identified the following domains: legal, medical, financial, Twitter posts, governmental, and oil and gas.

Figure~\ref{fig:filo} presents a phylogeny of language models developed for the Portuguese language, with the models identified in this study highlighted in bold. All identified models were developed from a preceding language model, with the following models serving as the basis: BERT~\citep{devlin2019bert}, T5~\citep{t5}, GPT~\citep{radford, radforda, radford2021learning}, Llama~\citep{llama1, llama2}, Mistral~\citep{jiang2023mistral}, Gemma~\citep{team2024gemma}, Phi~\citep{li2023textbooks, abdin2024phi3technicalreporthighly}, Qwen~\citep{qwen25}, and the multimodal models OpenCLIP~\citep{ilharco_2021_5143773} and FROMAGe~\citep{koh2023groundinglanguagemodelsimages}. These last two, in addition to using a language model as a basis, also utilize the ViT~\citep{dosovitskiy2020image} vision model.

The largest phylogeny of models identified in this study originated with BERT~\citep{devlin2019bert}, which serves as the foundation for 20 different models for the Portuguese language. The second-largest phylogeny was derived from the Llama family of models~\citep{llama1, llama2}, with 10 models for the Portuguese language. Furthermore, Figure~\ref{fig:filo} also shows information regarding the amount of derivations that some models underwent to become experts in their target tasks, for instance, the model derivation branch BERT~\citep{devlin2019bert} $\rightarrow$ RoBERTa~\citep{liu2019robertarobustlyoptimizedbert} $\rightarrow$   DeBERTa~\citep{he2020deberta} $\rightarrow$  Albertina-PT-*~\citep{albertina} $\rightarrow$   MediAlbertina-PT*~\citep{nunes_medialbertina_2024}. Another case shown in the Figure~\ref{fig:filo} is for the models which has more than one version, like the CardioBERTpt model~\citep{schneider_cardiobertpt_2023} that has 3 versions with different branches: the first one BERT~\citep{devlin2019bert} $\rightarrow$ mBERT~\citep{devlin2019bert} $\rightarrow$ BioBERTpt~\cite {schneider_biobertpt_2020}  $\rightarrow$ CardioBERTpt~\citep{schneider_cardiobertpt_2023}; the second one BERT~\citep{devlin2019bert} $\rightarrow$ mBERT~\citep{devlin2019bert} $\rightarrow$ CardioBERTpt~\citep{schneider_cardiobertpt_2023}; and the last one BERT~\citep{devlin2019bert} $\rightarrow$ BERTimbau~\citep{souza_bertimbau_2020} $\rightarrow$  CardioBERTpt~\citep{schneider_cardiobertpt_2023}.

\subsection{Availability of Models}
\label{sec:disponibilidade}

\begin{table}[p]
    \caption{Availability of language models for Portuguese.}
    \centering
    \renewcommand{\arraystretch}{1.1}
    \footnotesize \arrayrulecolor{black!5}
    \setlength{\tabcolsep}{2.25pt} 
    \begin{tabular}{>{\raggedright\arraybackslash}p{0.33\textwidth}|>{\centering\arraybackslash}p{0.099\textwidth}|>{\centering\arraybackslash}p{0.099\textwidth}|>{\centering\arraybackslash}p{0.099\textwidth}|>{\centering\arraybackslash}p{0.126\textwidth}|>{\centering\arraybackslash}p{0.126\textwidth}} \arrayrulecolor{black} \hline

    \textbf{Language Model} & \textbf{Code} & \textbf{Model} & \textbf{Data} & \textbf{Paper/Report} & \textbf{Model~Cards} \\\hline \arrayrulecolor{black!5}

      GPorTuguese-2\tiny~\citep{pierre2020gpt2smallportuguese} & \yt & \yt & \yt & \yt & \mt \\\hline
        
      BertPT, AlbertPT\tiny~\citep{feijo_mono_2020} & \yt & \yt & \mt & \yt & \nt \\\hline
        
      PTT5-v1\tiny~\citep{carmo_ptt5_2020} & \yt & \yt & \mt & \yt &  \mt \\\hline
        
      BERTimbau\tiny~\citep{souza_bertimbau_2020} & \nt & \yt & \mt & \yt & \mt \\\hline
        
      BioBERTpt\tiny~\citep{schneider_biobertpt_2020} & \nt & \yt & \nt & \yt & \mt \\\hline

      BERTaú\tiny~\citep{bertau} & \nt & \nt & \nt & \yt & \nt \\\hline

      GPT2-Bio-Pt\tiny~\citep{schneider_gpt-2_2021} & \yt & \yt & \yt & \yt & \mt \\\hline
        
      BERTikal\tiny~\citep{polo_legalnlp_2021} & \nt & \yt & \nt & \yt & \mt \\\hline
        
      PetroBERT\tiny~\citep{petrobert} & \nt & \nt & \mt & \yt & \nt \\\hline
      
      AiBERTa\tiny~\citep{miquelina_generating_2022} & \nt & \yt & \nt & \yt & \nt \\\hline
      
      JurisBERT\tiny~\citep{viegas_jurisbert_2023} & \yt & \yt & \yt & \yt & \mt \\\hline
      
      CardioBERTpt\tiny~\citep{schneider_cardiobertpt_2023} & \nt & \yt & \nt & \yt & \mt \\\hline
      
      FinBERT-PT-BR\tiny~\citep{santos_finbert-pt-br_2023} & \nt & \yt & \nt & \yt & \mt \\\hline
      
      Cabra\tiny~\citep{botbotroboticscabra_2024} & \nt & \yt & \yt & \nt & \mt \\\hline
      
      OpenCabrita\tiny~\citep{larcher_cabrita_2023} & \yt & \yt & \yt & \yt & \mt \\\hline
      
      BERTabaporu\tiny~\citep{da_costa_bertabaporu_2023} & \nt & \yt & \nt & \yt & \mt \\\hline
      
      DeBERTinha\tiny~\citep{debertinha} & \nt & \yt & \mt & \yt &  \mt \\\hline
      
      LegalBert-pt\tiny~\citep{silveira_legalbert} & \nt & \yt & \nt & \yt & \mt \\\hline
      
      Sabiá, Sabiá-J\tiny~\citep{pires_sabia_2023} & \nt & \mt & \mt &  \yt & \mt \\\hline
      
      Canarim-7B\tiny~\citep{maicon_domingues_2023} & \nt & \yt & \yt & \nt & \mt \\\hline
      
      CAPIVARA\tiny~\citep{dos_santos_capivara_2023} & \yt & \yt & \yt & \yt & \yt \\\hline
      
      Albertina PT-*\tiny~\citep{albertina} & \nt & \yt & \mt & \yt & \mt \\\hline
      
      Bode\tiny~\citep{garcia_introducing_2024} & \nt & \yt & \yt & \yt & \mt \\\hline

      Samba\tiny~\citep{samba} & \nt & \yt & \nt & \nt & \nt \\\hline
      
      GlórIA\tiny~\citep{lopes2024gloria} & \yt & \yt & \mt & \yt & \mt \\\hline
      
      PeLLE\tiny~\citep{demello2024pelle} & \nt & \nt & \yt & \yt & \nt \\\hline

      RoBERTaLexPT\tiny~\citep{garcia_robertalexpt_2024} & \nt & \yt & \yt & \yt & \mt \\\hline
            
      Sabiá-2\tiny~\citep{almeida_sabia-2_2024}  & \nt & \nt & \nt & \yt & \nt \\\hline
      
      Juru\tiny~\citep{junior2024juru} & \nt & \yt & \nt & \yt &  \mt \\\hline
      
      TeenyTinyLlama\tiny~\citep{teenytinyllama} & \yt & \yt & \yt & \yt & \yt \\\hline

      Albertina PT-*$^2$\tiny~\citep{santos-etal-2024-fostering} & \nt & \yt & \yt & \yt & \mt \\\hline
      
      Gervásio PT*\tiny~\citep{santos_advancing_2024} & \nt & \yt & \yt & \yt & \mt \\\hline
      
      MediAlbertina\tiny~\citep{nunes_medialbertina_2024} & \nt & \yt & \nt & \yt & \mt \\\hline
      
      Sabiá-3\tiny~\citep{abonizio_sabia-3_2025} & \nt & \nt & \nt & \yt & \nt \\\hline
      
      Tucano\tiny~\citep{correa_tucano_2024} & \yt & \yt & \yt & \yt & \yt \\\hline
      
      Carvalho\_pt-gl\tiny~\citep{gamallo_galician-portuguese_2025} & \nt & \yt & \mt & \yt & \mt \\\hline
      
      Serafim PT*\tiny~\citep{gomes_open_2025} & \nt & \yt & \yt & \yt & \mt  \\\hline
      
      V-GlórIA\tiny~\citep{simplicio_v-gloria_2024} & \yt & \mt & \mt & \yt & \nt \\\hline
      
      GemBode, PhiBode\tiny~\citep{garcia_gembode_2025} & \nt & \yt & \yt & \yt & \mt \\\hline
      
      BERTugues\tiny~\citep{zago_bertugues_2024} & \nt & \yt & \mt & \yt & \mt \\\hline
      
      BERTweet.BR\tiny~\citep{carneiro_bertweetbr_2025} & \nt & \yt & \nt & \yt & \mt \\\hline
      
      GovBERT-BR\tiny~\citep{silva_govbert-br_2025} & \nt & \yt & \mt & \yt  & \nt \\\hline
      
      PTT5-v2\tiny~\citep{piau_ptt5-v2_2025} & \nt & \yt & \yt & \yt & \mt \\\hline
      
      Clinical-BR-*\tiny~\citep{de_souza_pinto_developing_2025} & \nt & \yt & \mt & \yt & \mt \\\hline
      
      DeB3RTa\tiny~\citep{pires_deb3rta_2025} & \nt & \yt & \mt & \yt & \mt \\\hline
      
      Amadeus-Verbo\tiny~\citep{cruz-castaneda_amadeus-verbo_2025} & \nt & \yt & \nt & \yt & \mt \\\arrayrulecolor{black}\hline

    \end{tabular}
    \label{tab:dlmspt}

    \renewcommand{\arraystretch}{1.6}
    \scriptsize
    \textbf{Available:} \begin{tabular}{>{\centering\arraybackslash}p{0.09\textwidth}} \yt \end{tabular}
    \hspace{0.6cm}\textbf{Semi-available:} \begin{tabular}{>{\centering\arraybackslash}p{0.09\textwidth}} \mt \end{tabular}
    \hspace{0.6cm}\textbf{Not available:} \begin{tabular}{>{\centering\arraybackslash}p{0.09\textwidth}} \nt \end{tabular}
\end{table}

The availability of the language models developed for the Portuguese language identified in this study, separated by code, model, training data, and documentation via articles and Model Cards~\citep{mitchell2019}, is presented in Table~\ref{tab:dlmspt}. In this analysis, we considered the Model Cards, as HuggingFace currently provides a field on a model’s main page for filling out this documentation. Model Cards is an AI ethics tool~\citep{jhessicaspringer} designed to document a model's basic details of its creation and operation, intended use cases and users, and ethical considerations. 

Regarding code and model availability, we consider them available when they are open and accessible in a repository such as GitHub or HuggingFace. We consider a code and a model as semi-available (i) if there is an access link in some of the documentation, but it is necessary to request access authorization from the authors, or (ii) if the documentation introducing the model cites more than one version of that model, and not all are available for use. Finally, we consider a code and a model unavailable if there is no access link and no mention of availability in any documentation. Out of the 46 language models included in this study, only 11 made their code available~\citep{pierre2020gpt2smallportuguese, feijo_mono_2020, carmo_ptt5_2020, schneider_gpt-2_2021,  viegas_jurisbert_2023, larcher_cabrita_2023, dos_santos_capivara_2023, lopes2024gloria, teenytinyllama, correa_tucano_2024, garcia_gembode_2025}. In terms of model availability, 5 models are not available for use~\citep{bertau, petrobert, demello2024pelle, almeida_sabia-2_2024, abonizio_sabia-3_2025}, and 2 are partially available, where just the Sabiá-7B version is available for Sabiá models~\citep{pires_sabia_2023}, and the V-GlórIA model needs authorization~\citep{simplicio_v-gloria_2024}. Table~\ref{tab:hflmspt} provides links to access the models publicly available on HuggingFace or GitHub.

Regarding training data availability, we consider a dataset as available if it is open and can be downloaded, even if high computational resources are required for storage. We consider data as semi-available when either (i) the dataset is open, where there is an access link, but users must request permission from the authors to utilize it; or (ii) the original dataset is open, but the translated version, used for model training, has not been made available; or (iii) the model was trained using both open data that can be easily downloaded and closed/private data. Finally, we consider data as not available when it is closed and private and has not been made accessible to users. Out of the 46 language models included in this study, 17 were trained with available data~\citep{pierre2020gpt2smallportuguese, schneider_gpt-2_2021, viegas_jurisbert_2023, botbotroboticscabra_2024, larcher_cabrita_2023, maicon_domingues_2023, garcia_introducing_2024, demello2024pelle, garcia_robertalexpt_2024, teenytinyllama, santos-etal-2024-fostering, santos_advancing_2024, correa_tucano_2024, gomes_open_2025, garcia_gembode_2025, piau_ptt5-v2_2025, dos_santos_capivara_2023}, 15 were trained with non-available data~\citep{schneider_biobertpt_2020, samba, bertau, polo_legalnlp_2021, miquelina_generating_2022, schneider_cardiobertpt_2023, santos_finbert-pt-br_2023, da_costa_bertabaporu_2023, silveira_legalbert, almeida_sabia-2_2024, junior2024juru, nunes_medialbertina_2024, abonizio_sabia-3_2025, carneiro_bertweetbr_2025, cruz-castaneda_amadeus-verbo_2025}, and 14 with semi-available data, following the criteria (i), (ii), and (iii) mentioned above, (i)~\citep{souza_bertimbau_2020, pires_sabia_2023, albertina, de_souza_pinto_developing_2025, pires_deb3rta_2025, silva_govbert-br_2025, lopes2024gloria, feijo_mono_2020, zago_bertugues_2024, debertinha, carmo_ptt5_2020}, (ii)~\citep{simplicio_v-gloria_2024}, and (iii)~\citep{petrobert, gamallo_galician-portuguese_2025, lopes2024gloria}.

Regarding the availability of papers and reports, we consider a paper or report available if it is open access, whether on the original publication venue or on repositories such as arXiv, ResearchGate, or a blog-style page. We consider a paper or report as semi-available when it is not freely available and requires payment or a subscription to access it. Finally, we consider a paper or report as not available when there is no paper or report for that model, and only its HuggingFace page was found through the snowballing process. Out of the 46 language models included in this study, only Samba~\citep{samba}, Cabra~\citep{botbotroboticscabra_2024}, and Canarim-7B~\citep{maicon_domingues_2023} do not have a paper or a report. All the other models have a paper or report available\footnote{All papers were accessed through institutional access provided by the Universidade Estadual de Campinas.}. 

Regarding Model Card availability, which is currently covered by a dedicated page on HuggingFace, we consider a Model Card available if it includes all the sections proposed by~\citet{mitchell2019}, even if the information is organized differently. We consider a Model Card as semi-available when it has not been fully completed. Finally, we consider a Model Card as not available if none exists for that model. Out of the 46 language models included in this study, only CAPIVARA~\citep{dos_santos_capivara_2023}, TeenyTinyLlama~\citep{teenytinyllama}, and Tucano~\citep{correa_tucano_2024} had a complete Model Card. No Model Cards were found for 10 models~\citep{feijo_mono_2020, samba, bertau, petrobert, miquelina_generating_2022, demello2024pelle, almeida_sabia-2_2024, abonizio_sabia-3_2025, simplicio_v-gloria_2024, silva_govbert-br_2025}. The remaining analyzed models have semi-available Model Cards, primarily providing basic information about the model, the data, the metrics, the license, and the citation, and lacking details on factors, ethical considerations, or warnings, thereby confirming the quality assessment conducted in Section~\ref{sec:avql}.

\subsection{Training Data}
\label{sec:cdavaliacao}

We analyze the datasets used to train the identified language models for Portuguese, focusing on their domains, origins, and roles within different modeling strategies. To that end, we group datasets into categories based on their domain and data characteristics.

\textit{Web / Crawled Data} refers to large-scale corpora collected from the web, typically used in general-purpose models. \textit{Legal / Governmental} datasets include institutional and regulatory documents, while \textit{Clinical / Healthcare} datasets comprise medical and health-related data. \textit{Instruction / Conversational} datasets consist of instruction-response pairs used for fine-tuning language models. \textit{Multilingual / Parallel} datasets contain aligned text across multiple languages, commonly used for translation and cross-lingual tasks. The remaining categories, such as \textit{Financial / Economic}, \textit{News / Media}, and \textit{Scientific / Academic}, represent domain-specific corpora associated with particular application contexts.

\begin{figure}
    \centering
    \includegraphics[width=0.99\linewidth, trim={0cm, 7cm, 0cm, 6cm},clip]{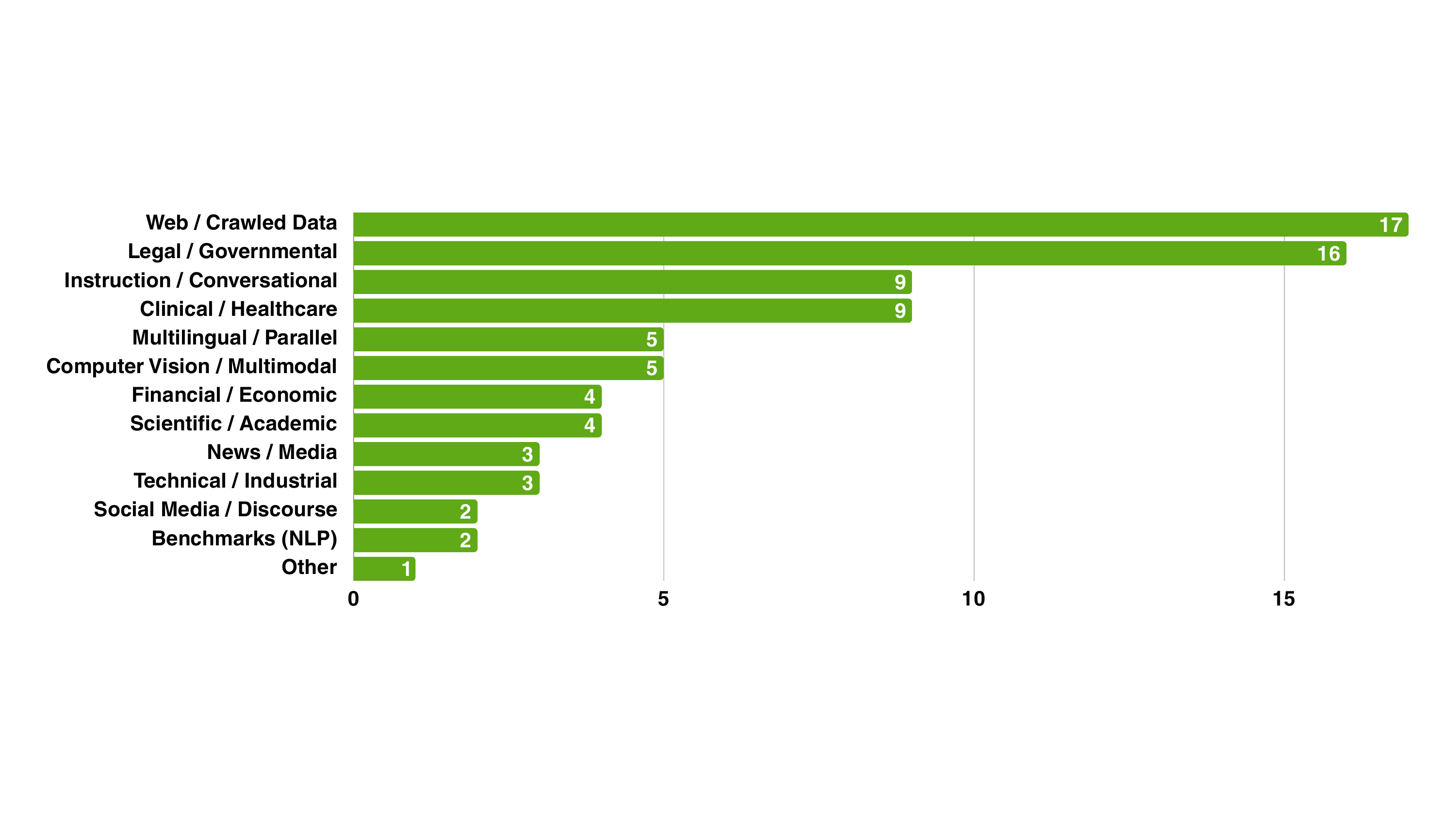}
    \caption{Overview of dataset categories used in language models for Portuguese. 
    }
    \label{fig:datasets}
\end{figure}

Figure~\ref{fig:datasets} presents the distribution of datasets across categories, highlighting the diversity of data sources employed across models. This diversity reflects the heterogeneity of the Portuguese NLP ecosystem, where different types of models rely on distinct data sources depending on their intended applications.

As shown in Figure~\ref{fig:datasets}, \textit{Web / Crawled Data} constitutes the largest category of datasets used to train language models for Portuguese, indicating a strong reliance on large-scale corpora collected from the web for general-purpose modeling. \textit{Legal / Governmental} datasets also represent a significant portion, reflecting the presence of models specialized in institutional and regulatory domains. In contrast, categories such as \textit{Clinical / Healthcare}, \textit{Instruction / Conversational}, and \textit{Multilingual / Parallel} appear in smaller proportions, suggesting that these types of data are typically associated with more targeted applications. The remaining categories, including \textit{Financial / Economic}, \textit{News / Media}, and \textit{Scientific / Academic}, are less represented, further highlighting the concentration of datasets in a few dominant sources.

These observations also reveal a clear relationship between the type of training data and the models' intended purpose. General-purpose language models~\citep{teenytinyllama, pierre2020gpt2smallportuguese, zago_bertugues_2024, pires_deb3rta_2025, feijo_mono_2020, correa_tucano_2024} tend to rely predominantly on \textit{Web / Crawled Data}, leveraging its scale and diversity to support a wide range of tasks. In contrast, domain-specific models are typically trained on curated datasets tailored to their respective application areas. For instance, models designed for legal applications, such as Juru~\citep{junior2024juru} and JurisBERT~\citep{viegas_jurisbert_2023}, make use of \textit{Legal / Governmental} corpora. Other examples are models in the healthcare domain, such as CardioBERTpt~\citep{schneider_cardiobertpt_2023}, MediAlbertina~\citep{nunes_medialbertina_2024}, and Clinical-BR~\citep{de_souza_pinto_developing_2025}, which rely on \textit{Clinical / Healthcare} data. This alignment between data source and model objective indicates that the choice of training data is largely driven by the target domain, rather than by a single standardized data collection strategy.

Another notable aspect is the presence of \textit{Instruction / Conversational} datasets, which are increasingly adopted in more recent models (e.g., Amadeus-Verbo~\citep{cruz-castaneda_amadeus-verbo_2025} and GemBode~\citep{garcia_gembode_2025}). These datasets, typically composed of instruction-response pairs, are used to fine-tune language models for interactive and task-oriented applications. Their inclusion reflects a shift in modeling practices, in which models are trained not only to generate coherent text but also to follow user instructions more effectively. Although smaller in scale than web-crawled corpora, instruction-based datasets play a key role in adapting general-purpose models to more practical, user-centered scenarios~\citep{teenytinyllama}.

Overall, the distribution of datasets highlights the heterogeneity of Portuguese NLP models, with multiple data sources and strategies coexisting. Rather than relying on a single standardized corpus, model development follows different approaches depending on the intended application, data availability, and domain requirements. This diversity of datasets reflects both the opportunities and challenges associated with building language models for Portuguese, as it enables the development of specialized systems while also indicating that such models are often trained on multiple heterogeneous sources rather than a single large-scale standardized corpus.

\section{Gaps and Opportunities}
\label{sec:gaps}

This section highlights the main gaps and opportunities in language models for Portuguese. We divide the discussion into the following topics: The Portuguese language and the lack of representation of its diversity (Section~\ref{sec:pt-diversity});  Use of translated data and the quality of the data used (Section~\ref{sec:data-quality}); Lack of Qualitative and Ethical Analysis (Section~\ref{sec:pt-qualitativeEthics}); Model Availability and Access Methods (Section~\ref{sec:pt-avail-acess}); and Multimodal models (Section~\ref{sec:multimodal}).

\subsection{The Portuguese Language and the Lack of Representation of its Diversity}
\label{sec:pt-diversity}

In this systematic mapping study, we found only language models designed for Brazilian Portuguese, European Portuguese, or both variants. An important point is that when a model was trained on textual data from both variants, some authors referred to it simply as a ``Portuguese model'', assuming that it could be applied and used in all scenarios involving the Portuguese language. However, assuming that a model trained with textual data in Brazilian and European Portuguese is sufficient to serve Portuguese-speaking populations is a linguistic prejudice that needs to be addressed.

The Portuguese language is pluricentric, being the official language in nine countries: Angola, Brazil, Cape Verde, Guinea-Bissau, Equatorial Guinea, Mozambique, Portugal, São Tomé and Príncipe, East Timor, all members of The Community of Portuguese Speaking Countries (CPLP -- \textit{Comunidade dos Países de Língua Portuguesa})\footnote{\url{https://www.cplp.org/}}, and in an administrative region of China, Macau. The Portuguese language is present in four continents (the Americas, Africa, Europe, and Asia),  has more than 250 million speakers, and is the sixth most-spoken language on the internet~\citep{pimientaorganizaccao}. Although these Portuguese-speaking countries share a common linguistic foundation, each has adapted the language to its cultural, historical, social, and economic contexts, thereby creating its own norms governing language use~\citep{da2026multilinguismo}.  If we consider Brazil, the researchers' country, the Portuguese language has undergone numerous transformations influenced by both Indigenous and African languages. Additionally, because of its vast territory, Brazil possesses a great complexity of idiomatic and regional expressions, different accents, and ways of speaking. Such complexities may not be adequately represented by the models designed for Brazilian Portuguese.

Therefore, the community conducting research and developing language models for Portuguese must consider this diversity across Portuguese-speaking countries, combating linguistic prejudice by recognizing that a model trained solely on textual data from Brazilian Portuguese combined with textual data from European Portuguese cannot possibly suffice as a model for general Portuguese. As future research directions, we see an opportunity to create models more capable of addressing the specificities of each Portuguese variant, which will require efforts to curate datasets that represents the cultural, historical, social, and economic aspects of the countries where the models will be deployed.

\subsection{Use of Translated Data and the Quality of the Data Used }
\label{sec:data-quality}

As shown in Section~\ref{sec:over} (see Table~\ref{tab:mi2}), some language models for Portuguese were trained using text data translated from English to Portuguese via machine translation. Consequently, even though these models are intended for Portuguese-speaking populations, they tend to replicate aspects of the global North that do not represent these populations, and they also carry errors stemming from machine translations~\citep{johnson2026ghost, dos_santos_capivara_2023, hovy2021five}. Moreover, because they have not been exposed to original Portuguese data capturing the sociocultural aspects of Portuguese-speaking countries, such models may not perform well on tasks that require this knowledge. 

One of the main arguments made by papers proposing models for Portuguese to justify the use of English-to-Portuguese text data is that the model was designed to fill the gap of serving a language with limited computational resources in terms of data ready for AI training. However, through this mapping, we found that over the years (between 2020 and 2025), new corpora and datasets for training language models in Portuguese have been created, particularly for the Brazilian variant, such as Jabuticaba~\citep{Jabuticaba} and Carolina~\citep{crespo2023carolinageneralcorpuscontemporary}. 
Therefore, future work on developing models for the Portuguese language needs to reconsider the use of translated data and invest in creating and using well-curated, high-quality corpora for the Portuguese variants. Initiatives such as the one by The International Portuguese Language Institute (IILP -- \textit{Instituto Internacional de Língua Portuguesa}) and Centre of Linguistics of the University of Porto (CLUP -- \textit{Centro de Linguística da Universidade do Porto}), which proposes the creation of an oral corpus for the Portuguese language while considering all variants\footnote{\url{https://iilp.cplp.org/corpus-portugues-oral-cplp/}}, are good examples of direction for future work.

\subsection{Lack of Qualitative and Ethical Analysis}
\label{sec:pt-qualitativeEthics}

It is important to revisit the analyses conducted in Section~\ref{sec:avql}. We found that few papers introducing language models for Portuguese have conducted ethical and qualitative analyses of the proposed models. From the 46 mapped models, only 5 included a discussion of ethical issues in the papers introducing them, 17 addressed the model’s limitations, and 8 conducted a qualitative analysis. 
Although all models presented some form of quantitative analysis based on a metric defined in the literature, the community must, when proposing such models, not only consider the metric performance but also consider qualitative analyses of the responses returned by the model, since we cannot affirm that a model attends a language and a population if qualitative and ethical analyses are not conducted with care.

Currently, there are practical tools that can assist in ethical and qualitative analyses of AI models~\citep{jhessicaspringer}, such as Model Cards~\citep{mitchell2019}, which proposes the creation of more responsible and transparent documentation about the models and is already incorporated into the default HuggingFace model page --- the most common platform where such models are hosted, as shown in Table~\ref{tab:hflmspt}. However, of the 46 mapped models, only 3 had a complete Model Cards documentation~\citep{dos_santos_capivara_2023, teenytinyllama, correa_tucano_2024}. Responsible documentation of a model should no longer be viewed as an option, and models must be accompanied by documentations describing why the model was created, its intended uses, who the model is intended for, and who is underrepresented, as well as the model’s social, cultural, economic, and environmental impacts. Regarding this last point, some papers introducing models to the Portuguese-speaking population provided an approximate value for the model’s carbon footprint. However, we note that the papers provide measurement only for the final version of the model and without a robust method.  Thus, it is important to make available the carbon footprint of the models in this documentation, taking into account the entire model development cycle.

\subsection{Model Availability and Access Methods}
\label{sec:pt-avail-acess}

As noted in Section~\ref{sec:disponibilidade} (see  Table~\ref{tab:dlmspt}), from the 46 mapped models, only 11 had their source code available. Without making the source code available, the model development it is not fully transparent, which can hinder the reproducibility of reported results and limit progress in the field. Furthermore, it is important to highlight that, although most models are available for community use via HuggingFace or GitHub (see Table~\ref{tab:hflmspt}), only individuals with technical expertise actually have access to the models, since it is necessary to know how to execute it with code. Thus, based on the analyzed documentation of the mapped models, we observed that most models are not available to end users through user-friendly interaction systems. Currently, there are systems that use some of the mapped models in this study as a backend, such as the chatbots Maritaca Chat\footnote{\url{https://chat.maritaca.ai/auth}}, which uses models from the Sabiá family~\citep{pires_sabia_2023, almeida_sabia-2_2024, abonizio_sabia-3_2025},  Evaristo.ai\footnote{\url{https://evaristo.ai/}}, which uses the Gervásio PT* model~\citep{santos_advancing_2024}, and TeenyTinyLlama-Chat\footnote{\url{https://huggingface.co/spaces/nicholasKluge/TeenyTinyLlama-Chat}}, which uses the TeenyTinyLlama model~\citep{teenytinyllama}. Thus, we see the availability of these models as an opportunity not only for researchers and professionals in the field but also for the general public. However, we emphasize, as discussed earlier, that such advances require better curation of the models, both qualitatively and ethically, so that they can be delivered to end users with improved analyses of their functioning and better documentation.

\subsection{Multimodal models}
\label{sec:multimodal}

In this systematic mapping study, we included terms in our search string to also identify multimodal models focused on the Portuguese language (\textit{i.e.}, ``vision-language'', ``audio-language'', and ``multimodal''). Additionally, the third control paper used to calibrate the search string was itself a multimodal model (CAPIVARA~\citep{dos_santos_capivara_2023}). However, our automated search returned only two vision-language models~\citep{dos_santos_capivara_2023, simplicio_v-gloria_2024}, and both are general in vision language tasks. Therefore, the first opportunity for multimodal models we envision is the development of models targeting other modalities (\textit{e.g.}, audio) and specific to domains such as legal or medical ones. 

Following trends observed for English-based multimodal models, we identify two key avenues for future contributions: (1) extending the models to three or more modalities, as in the LanguageBind~\citep{zhu2024languagebind} and ImageBind~\citep{girdhar2023imagebind} models; and (2) leveraging advanced LLMs as a central backbone of a multimodal architecture along with modality-to-text and text-to-modality projectors, yielding the so-called multimodal large language models (MLLM~\citep{caffagni2024revolution} or MM-LLM~\citep{zhang2024mm}). Existing multimodal datasets comprising non-translated Portuguese content like \#\textit{PraCegoVer}~\citep{PraCegoVer}, Framed Multi30k~\citep{viridiano2024framed}, and pt-image-ir-dataset~\citep{duarte2026pt} may support this development. However, considering the growing research efforts to advance multimodal models, proposing new corpora is essential to address the current lack of such resources.

As these models involve text, they are susceptible to the same issues as LLMs (Subsections \ref{sec:pt-diversity} and \ref{sec:data-quality}). In fact, the lack of regional representation affects both text and images, as visual elements common in one region may not be representative of another. Furthermore, existing studies generally rely on images from English-centered datasets paired with translated versions of the texts. In this sense, in addition to diverse textual data, regionally- or culturally-driven images~\citep{romero2024cvqa} also represent a promising direction for future research.

\section{Conclusion}
\label{sec:conclusion}

In this systematic mapping study, we present an analysis of language models developed for the Portuguese language, providing a comprehensive overview of the current state of the field. To this end, we considered models published between January 2020 and August 2025, indexed in Scopus, IEEEXplore, Web of Science, and arXiv, and supplemented the search with forward and backward snowballing. We filtered all the identified studies according to inclusion and exclusion criteria, resulting in 46 language models for Portuguese. Our research focused on categorizing the mapped models based on aspects such as base model, architecture, computational resources, training datasets, licensing, and code, data, and model weights availability. In addition, we analyzed the evolution and relationships among these models adopting a phylogenetic perspective and assessed the quality of the accompanying documentations.

We identified several limitations and challenges in existing language models for Portuguese, including i) studies that reduce the Portuguese language to only its Brazilian and European variants, without considering the diversity of the language and its variations; ii) the use of translated data that does not represent the cultural, historical, social, and economic aspects of the countries where the models will be deployed; iii) a lack of qualitative and ethical analysis, and little discussion of the models’ limitations; iv) limited reproducibility due to the lack of open-source code; v) the unavailability of the models to end users through user-friendly interaction systems; and vi) a lack of multimodal models for the Portuguese language.

To address these limitations, future studies should focus on i) developing models that are tailored to the characteristics of each Portuguese variant, which will require efforts to curate datasets that reflect aspects of the countries where the models will be deployed; ii) conducting qualitative and ethical analyses of the models, as we cannot affirm that a model adequately serves a language and a population without careful examination; iii) advacing multimodal models that incorporate the Portuguese language; and iv) sharing these models, not only with researchers and professionals in the field but also with the general public.

We hope that all these points will contribute to the further development of language models for Portuguese and spark the community’s interest in advancing models tailored to Portuguese speakers.

\section{Acknowledgments}

This project was supported by the Ministry of Science, Technology, and Innovation of Brazil, with resources granted by the Federal Law 8.248 of October 23, 1991, under the PPI-Softex. The project was coordinated by Softex and published as Intelligent agents for mobile platforms based on Cognitive Architecture technology [01245.003479/2024-10]. J.~Silva is funded by FAPESP 2024/23118-1 and was partially financed by the Coordination for the Improvement of Higher Education Personnel (CAPES) -- Finance Code 001. S.~Avila is also partially funded by FAPESP 2023/12086-9, 2023/12865-8, \mbox{2020/09838-0}, 2013/08293-7, and CNPq 316489/2023-9. H.~Pedrini is also partially funded by CNPq 304836/2022-2.

\newpage

\appendix
\section{Access to Language Models for Portuguese}
\label{sec:linkssec}

{\footnotesize
\renewcommand{\arraystretch}{1.5}
\arrayrulecolor{black!5}

\begin{longtable}{>{\raggedright\arraybackslash}p{0.28\textwidth}>{\raggedright\arraybackslash}p{0.65\textwidth}} 
    \caption{Access to language models for Portuguese. Accessed in April 2026.}
    \\\arrayrulecolor{black}\hline

    \textbf{Language Model} & \textbf{HuggingFace/GitHub} \\\hline \arrayrulecolor{black!5}

      GPorTuguese-2\tiny~\citep{pierre2020gpt2smallportuguese} & \scriptsize\url{https://huggingface.co/pierreguillou/gpt2-small-portuguese} \\\hline
      
      BertPT, AlbertPT\tiny~\citep{feijo_mono_2020} & \scriptsize\url{https://github.com/diego-feijo/bertpt/} \\\hline
      
      PTT5-v1\tiny~\citep{carmo_ptt5_2020} & \scriptsize\url{https://huggingface.co/collections/unicamp-dl/ptt5} \\\hline
      
      BERTimbau\tiny~\citep{souza_bertimbau_2020} & \scriptsize\url{https://huggingface.co/neuralmind/bert-large-portuguese-cased} \hspace{0.2cm}and\hspace{0.2cm}  \scriptsize\url{https://huggingface.co/neuralmind/bert-base-portuguese-cased} \\\hline
      
      BioBERTpt\tiny~\citep{schneider_biobertpt_2020} & \scriptsize\url{https://huggingface.co/pucpr/biobertpt-all} \\\hline

      BERTaú\tiny~\citep{bertau} & Not publicly available. \\\hline
      
      GPT2-Bio-Pt\tiny~\citep{schneider_gpt-2_2021} & \scriptsize\url{https://huggingface.co/pucpr/gpt2-bio-pt} \\\hline
      
      BERTikal\tiny~\citep{polo_legalnlp_2021} & \scriptsize\url{https://huggingface.co/felipemaiapolo/legalnlp-bert} \\\hline
      
      PetroBERT\tiny~\citep{petrobert} & Not publicly available. \\\hline
      
      AiBERTa\tiny~\citep{miquelina_generating_2022} & \scriptsize\url{https://huggingface.co/AiBERTa} \\\hline
      
      JurisBERT\tiny~\citep{viegas_jurisbert_2023} & \scriptsize\url{https://huggingface.co/alfaneo/jurisbert-base-portuguese-uncased} \\\hline
      
      CardioBERTpt\tiny~\citep{schneider_cardiobertpt_2023} & \scriptsize\url{https://huggingface.co/pucpr-br/cardiobertpt} \\\hline
      
      FinBERT-PT-BR\tiny~\citep{santos_finbert-pt-br_2023} & \scriptsize\url{https://huggingface.co/lucas-leme/FinBERT-PT-BR} \\\hline
      
      Cabra\tiny~\citep{botbotroboticscabra_2024} & \scriptsize\url{https://huggingface.co/botbotrobotics/Cabra} \\\hline
      
      OpenCabrita\tiny~\citep{larcher_cabrita_2023} & \scriptsize\url{https://huggingface.co/22h/open-cabrita3b} \\\hline
      
      BERTabaporu\tiny~\citep{da_costa_bertabaporu_2023} & \scriptsize\url{https://huggingface.co/pablocosta/bertabaporu-base-uncased} \hspace{0.2cm}and\hspace{0.2cm} \scriptsize\url{https://huggingface.co/pablocosta/bertabaporu-large-uncased} \\\hline
      
      DeBERTinha\tiny~\citep{debertinha} & \scriptsize\url{https://huggingface.co/collections/sagui-nlp/debertinha} \\\hline
      
      LegalBert-pt\tiny~\citep{silveira_legalbert} & \scriptsize\url{https://huggingface.co/raquelsilveira/legalbertpt_fp} \hspace{0.2cm}and\hspace{0.2cm}  \scriptsize\url{https://huggingface.co/raquelsilveira/legalbertpt_sc}\\\hline
      
      Sabiá \tiny~\citep{pires_sabia_2023} & Only 7b: \scriptsize\url{https://huggingface.co/maritaca-ai/sabia-7b} \\\hline
      
      Canarim-7B\tiny~\citep{maicon_domingues_2023} & \scriptsize\url{https://huggingface.co/dominguesm/canarim-7b} \\\hline
      
      CAPIVARA\tiny~\citep{dos_santos_capivara_2023} & \scriptsize\url{https://huggingface.co/hiaac-nlp/CAPIVARA} \hspace{0.2cm}and\hspace{0.2cm}  \scriptsize\url{https://huggingface.co/hiaac-nlp/CAPIVARA-Opt}\\\hline
      
      Albertina PT-*\tiny~\citep{albertina, santos-etal-2024-fostering} & \scriptsize\url{https://huggingface.co/collections/PORTULAN/albertina}\\\hline
      
      Bode\tiny~\citep{garcia_introducing_2024} & \scriptsize\url{https://huggingface.co/collections/recogna-nlp/bode-llm-em-portugues} \\\hline

      Samba\tiny~\citep{samba} & \scriptsize\url{https://huggingface.co/lrds-code/samba-1.1B} \\\hline
      
      GlórIA\tiny~\citep{lopes2024gloria} & \scriptsize\url{https://huggingface.co/NOVA-vision-language/GlorIA-1.3B} \\\hline
      
      PeLLE\tiny~\citep{demello2024pelle} & Not publicly available. \\\hline

      RoBERTaLexPT\tiny~\citep{garcia_robertalexpt_2024} & \scriptsize\url{https://huggingface.co/eduagarcia/RoBERTaLexPT-base} \hspace{0.2cm}and\hspace{0.2cm} \scriptsize\url{https://huggingface.co/eduagarcia/RoBERTaCrawlPT-base}\\\hline
      
      Sabiá-2\tiny~\citep{almeida_sabia-2_2024} & Not publicly available. \\\hline
      
      Juru\tiny~\citep{junior2024juru} & \scriptsize\url{https://huggingface.co/roseval/Juru-7B} \\\hline
      
      TeenyTinyLlama\tiny~\citep{teenytinyllama} & \scriptsize\url{https://huggingface.co/collections/nicholasKluge/teenytinyllama} \\\hline
      
      Gervásio PT*\tiny~\citep{santos_advancing_2024} & \scriptsize\url{https://huggingface.co/collections/PORTULAN/gervasio}\\\hline
      
      MediAlbertina\tiny~\citep{nunes_medialbertina_2024} & \scriptsize\url{https://huggingface.co/portugueseNLP/medialbertina_pt-pt_900m} \hspace{0.2cm}and\hspace{0.2cm} \scriptsize\url{https://huggingface.co/portugueseNLP/medialbertina_pt-pt_1.5b} \\\hline
      
      Sabiá-3\tiny~\citep{abonizio_sabia-3_2025} & Not publicly available. \\\hline
      
      Tucano\tiny~\citep{correa_tucano_2024} & \scriptsize\url{https://huggingface.co/collections/TucanoBR/tucano} \\\hline
      
      Carvalho\_pt-gl\tiny~\citep{gamallo_galician-portuguese_2025} & \scriptsize\url{https://huggingface.co/collections/Nos-PT/carvalho-family-67e423bf209c732396377b61}\\\hline
      
      Serafim PT*\tiny~\citep{gomes_open_2025}  & \scriptsize\url{https://huggingface.co/collections/PORTULAN/serafim-66a39e91d73bea6c16105873}\\\hline
      
      V-GlórIA\tiny~\citep{simplicio_v-gloria_2024} &  \scriptsize\url{https://github.com/amsimplicio/V-GlorIA?tab=readme-ov-file} \\\hline
      
      GemBode, PhiBode\tiny~\citep{garcia_gembode_2025}  & \scriptsize\url{https://huggingface.co/collections/recogna-nlp/gembode} \hspace{0.2cm}and\hspace{0.2cm} \scriptsize\url{https://huggingface.co/collections/recogna-nlp/phibode}\\\hline
      
      BERTugues\tiny~\citep{zago_bertugues_2024} & \scriptsize\url{https://huggingface.co/ricardoz/BERTugues-base-portuguese-cased} \\\hline
      
      BERTweet.BR\tiny~\citep{carneiro_bertweetbr_2025}  & \scriptsize\url{https://huggingface.co/melll-uff/bertweetbr}\\\hline
      
      GovBERT-BR\tiny~\citep{silva_govbert-br_2025} & \scriptsize\url{https://huggingface.co/dccmpmgfinalisticas/GovBERT-BR} \\\hline
      
      PTT5-v2\tiny~\citep{piau_ptt5-v2_2025} & \scriptsize\url{https://huggingface.co/collections/unicamp-dl/ptt5-v2} \\\hline
      
      Clinical-BR-*\tiny~\citep{de_souza_pinto_developing_2025} & \scriptsize\url{https://huggingface.co/pucpr-br/Clinical-BR-Mistral-7B-v0.2} \hspace{0.2cm}and\hspace{0.2cm}  \scriptsize\url{https://huggingface.co/pucpr-br/Clinical-BR-LlaMA-2-7B}\\\hline
      
      DeB3RTa\tiny~\citep{pires_deb3rta_2025} & \scriptsize\url{https://huggingface.co/higopires/DeB3RTa-base} \hspace{0.2cm}and\hspace{0.2cm}  \scriptsize\url{https://huggingface.co/higopires/DeB3RTa-small}\\\hline
      
      Amadeus-Verbo\tiny~\citep{cruz-castaneda_amadeus-verbo_2025} & \scriptsize\url{https://huggingface.co/collections/amadeusai/amadeus-verbo-qwen25-pt-br-powered-by-aws} \\\arrayrulecolor{black}\hline
        
    \label{tab:hflmspt}
\end{longtable}
}
\section{Authors' Affiliations Acronyms List}
\label{sec:acro}

{\footnotesize
\begin{itemize}
\item \textbf{22h -}  22h
\item \textbf{Alfaneo -}	Alfaneo
\item \textbf{Amadeus AI -}	Amadeus AI
\item \textbf{BotBot -}	BotBot
\item \textbf{Comsentimento -}	Comsentimento
\item \textbf{Datalab -}	Latam Datalab Serasa Experian
\item \textbf{FEI -}	Centro Universitário da Fundação Educacional Inaciana
\item \textbf{HES-SO -}	University of Applied Sciences and Arts of Western Switzerland
\item \textbf{HiMolde -}	Molde University College
\item \textbf{IBM Research -}	IBM Research
\item \textbf{IFCE -}	Instituto Federal do Ceará
\item \textbf{IFF -}	Instituto Federal Fluminense
\item \textbf{IFMA -}	Instituto Federal do Maranhão
\item \textbf{ISCTE -}	Instituto Universitário de Lisboa
\item \textbf{Itaú Unibanco -}	Itaú Unibanco
\item \textbf{JusBrasil -}	JusBrasil
\item \textbf{Maritaca AI -}	Maritaca AI
\item \textbf{NeuralMind -}	NeuralMind 
\item \textbf{NOVA LINCS -}	NOVA Laboratory for Computer
Science and Informatics 
\item \textbf{PUC-PR -}	Pontifícia Universidade Católica Do Paraná
\item \textbf{PUC-Rio -}	Pontifícia Universidade Católica do Rio de Janeiro
\item \textbf{PUC-RS -}	Pontifícia Universidade Católica do Rio Grande do Sul
\item \textbf{Sagui AI -}	Sagui AI
\item \textbf{Select Data -}	Select Data
\item \textbf{Tikal Tech -}	Tikal Tech
\item \textbf{U.Porto -}	Universidade do Porto
\item \textbf{UÉ -}	Universidade de Evora
\item \textbf{UFF -}	Universidade Federal Fluminense
\item \textbf{UFG -}	Universidade Federal de Goiás
\item \textbf{UFMA -}	Universidade Federal do Maranhão
\item \textbf{UFMG -}	Universidade Federal de Minas Gerais
\item \textbf{UFMS -}	Universidade Federal de Mato Grosso do Sul
\item \textbf{UFRGS -}	Universidade Federal do Rio Grande do Sul
\item \textbf{UFRPE -}	Universidade Federal Rural de Pernambuco
\item \textbf{ULisboa -}	Universidade de Lisboa
\item \textbf{UMich -}	University of Michigan
\item \textbf{UNESP -}	Universidade Estadual de São Paulo
\item \textbf{Uni Bonn -}	University of Bonn
\item \textbf{UNICAMP -}	Universidade Estadual de Campinas
\item \textbf{UNIFOR -}	Universidade de Fortaleza
\item \textbf{UPF -}	Universitat Pompeu Fabra
\item \textbf{USC -}	Universidade de Santiago de Compostela
\item \textbf{USP -}	Universidade de São Paulo 
\item \textbf{UTFPR -}	Universidade Tecnológica Federal do Paraná
\item \textbf{UWaterloo -}	University of Waterloo
\item \textbf{UnB -}    Universidade de Brasília	
\item \textbf{UNIGE -} University of Geneva	
\item \textbf{UFPE -} Universidade Federal de Pernambuco	
\end{itemize}
}



\bibliographystyle{abbrvnat}
\bibliography{references}

\label{lastpage}

\end{document}